\documentstyle[fullname,lingmacros,leqno]{article}
\author{Jan Jaspars\\
CWI\\ P.O. Box 94079,\\ 1090 GB Amsterdam,\\ The Netherlands\\
{\tt jaspars@cwi.nl}
\and
Megumi Kameyama\\
SRI International\\ 333 Ravenswood Ave,\\ Menlo Park,\\ CA 94025, U.S.A.\\
{\tt megumi@ai.sri.com}
}
\title{Discourse Preferences in Dynamic Logic%
\thanks{The first author's work was supported by CEC project
LRE-62-051 (FraCaS). The second author's work was in part supported by
the National Science Foundation and the Advanced Research Projects
Agency under Grant IRI--9314961 (Integrated Techniques for Generation
and Interpretation).  We would like to thank the two anonymous
reviewers for helpful comments on an earlier version of the paper.  }}
\date{}

\begin{document}
\bibliographystyle{fullname}
\maketitle
%Up and down-operators%%%%%%%%%%%%%%%%%%%%%%%%%%%%%%%%%%%%%
\newcommand{\upb}[1]{[ \hspace*{1.5pt} #1 \hspace*{1.5pt}]^{+}\hspace*{1.5pt}}
\newcommand{\upd}[1]{\langle \hspace*{1.5pt} #1\hspace*{1.5pt}\rangle^{+} \hspace*{1.5pt}}
\newcommand{\dob}[1]{[ \hspace*{1.5pt} #1 \hspace*{1.5pt}]^{-}\hspace*{1.5pt}}
\newcommand{\dod}[1]{\langle \hspace*{1.5pt} #1\hspace*{1.5pt} \rangle^{-}\hspace*{1.5pt}}
%Update and downdate operators (minimal)%%%%%%%%%%%%%%%%%%%%
\newcommand{\upbm}[1]{[ \hspace*{1.5pt} #1\hspace*{1.5pt} ]^{+\mu}\hspace*{1.5pt}}
\newcommand{\updm}[1]{\langle \hspace*{1.5pt} #1\hspace*{1.5pt}\rangle^{+\mu} \hspace*{1.5pt}}
\newcommand{\dobm}[1]{[ \hspace*{1.5pt} #1\hspace*{1.5pt} ]^{-\mu}\hspace*{1.5pt}}
\newcommand{\dodm}[1]{\langle \hspace*{1.5pt} #1\hspace*{1.5pt} \rangle^{-\mu}\hspace*{1.5pt}}
%Theorems%%%%%%%%%%%%%%%%%%%%%%%%%%%%%%%%%%%%%%%%%%%%%%%%%%
\newtheorem{defi}{Definition}
\newtheorem{exa}{Example}
%Environments%%%%%%%%%%%%%%%%%%%%%%%%%%%%%%%%%%%%%%%%%%%%%%
\newenvironment{our-definition}{\begin{defi}\rm}{\end{defi}}
\newenvironment{our-example}{\begin{exa}\rm}{\end{exa}}
%Symbols%%%%%%%%%%%%%%%%%%%%%%%%%%%%%%%%%%%%%%%%%%%%%%%%%%
\def\lll{\langle\!\langle}
\def\Bbb#1{I\!\!#1}
\newcommand{\rrr}{\rangle\!\rangle}
\newcommand{\sqq}{\sqsubseteq}
\newcommand{\ldb}{\lbrack\!\lbrack}
\newcommand{\rdb}{\rbrack\!\rbrack}
\newcommand{\pref}[1]{\ldb #1 \rdb_{\mbox{\scriptsize \sfp}}}
\newcommand{\lora}{\longrightarrow}
\newcommand{\Lolra}{\Longleftrightarrow}
\newcommand{\Lra}{\Leftrightarrow}
\newcommand{\Lora}{\Longrightarrow}
\newcommand{\Ra}{\Rightarrow}
\newcommand{\la}{\leftarrow}
\newcommand{\ra}{\rightarrow}
\newcommand{\A}{\forall}
\newcommand{\seq}{\subseteq}
\newcommand{\prr}{\mathop{\mbox{\sf p}}}
\newcommand{\pri}{\mathop{\mbox{\sf p}_i}}
\newcommand{\phrase}[2]{\begin{equation} \label{#2}\mbox{\it #1}\end{equation}}
\newcommand{\ld}[1]{{}^{#1}\ldb}
\newcommand{\M}[1]{M_{\mbox{\sf \scriptsize #1}}}
\newcommand{\Lp}{{\cal L}_{\mbox{\sf \scriptsize p}}}
\newcommand{\Lpm}{{\cal L}_{\mbox{\sf \scriptsize p},m}}\def\vast{\mathrel{\mkern-4mu}}\def\nmm{\mathrel|\vast\mathrel\approx}
\newcommand{\bfs}[1]{\mbox{\scriptsize\bf #1}}
\newcommand{\tts}[2]{{\sf #1}$_{\mbox{\scriptsize #2}}$}

%Figures%%%%%%%%%%%%%%%%%%%%%%%%%%%%%%%%%%%%%%%%%%%%%%%%%%%%%%%%%%%%
\newcommand{\fig}[2]{ \epsfxsize #2 \epsffile{#1}}

%MK's shorthands%%%%%%%%%%%%%%%%%%%%%%%%%%%%%%%%%%%%%%%%%%%%%%%%%%
\newcommand{\bit}{\begin{itemize}}
\newcommand{\eit}{\end{itemize}}
\newcommand{\ben}{\begin{enumerate}}
\newcommand{\een}{\end{enumerate}}
\newcommand{\bqt}{\begin{quote}}
\newcommand{\eqt}{\end{quote}}
\newcommand{\bc}{\begin{center}}
\newcommand{\ec}{\end{center}}
\newcommand{\bdes}{\begin{description}}
\newcommand{\edes}{\end{description}}
\newcommand{\btable}{\begin{footnotesize}\begin{table}}
\newcommand{\etable}{\end{table}\end{footnotesize}}
\newcommand{\bpack}{\begin{list}{$\bullet$}{\parsep 0pt \itemsep 4pt \topsep 4pt \parskip 0pt \partopsep 0pt \leftmargin 28pt}}
\newcommand{\epack}{\end{list}}

%%%%% macros for AC10, JJ: Feb. 11%%%%%

\def\At{\mbox{\sf Atoms}}
\def\Pred{\mbox{\sf Pred}}
\def\Con{\mbox{\sf Con}}
\def\Var{\mbox{\sf Var}}
\def\Dom{\mbox{\sf Dom}}
\def\bird{\mbox{\sl bird}}
\def\penguin{\mbox{\sl penguin}}
\def\fly{\mbox{\sl can--fly}}
\def\quaker{\mbox{\sl quaker}}
\def\republican{\mbox{\sl republican}}
\def\pacifist{\mbox{\sl pacifist}}
\def\Greet{\mathop{\mbox{\sl Greet}}}
\def\Meet{\mathop{\mbox{\sl Meet}}}
\def\Hit{\mathop{\mbox{\sl Hit}}}
\def\Injured{\mathop{\mbox{\sl Injured}}}
\def\john{\mbox{\sf j}} 
\def\bill{\mbox{\sf b}} 
\def\schw{\mbox{\sf s}\!\mbox{\sf c}\!\mbox{\sf h}} 
\def\cyb{\mbox{\sf t}\!\mbox{\sf m}}

\def\Prop{I\!\! P}
\def\L{I\!\! L}
\def\spdl{{\bf spdl}$_{\mbox{\scriptsize \bf 1}}$}
\def\pr#1{\mathop{\mbox{\sf p}_{#1}}}
\def\drs#1#2{\framebox{$%
               \begin{array}{l} #1 \\  \hline
               \multicolumn{1}{c}{#2}
               \end{array}$}}

\newtheorem{prc}{\sc Principle}
\newenvironment{principle}{\begin{prc} \rm}{\end{prc}}

\def\ignore#1{{}}
\def\phi{\varphi}
\def\rf#1{(\ref{#1})}
%%%%%%%%%%%%%%%%%%%%%%%%%%%%%%%%%%%%%%%%%%%%%%%%%%%%%%%%%%%%%%%%%%%%%%
%%% Abstract here
\begin{abstract}
In order to enrich dynamic semantic theories with a `pragmatic'
capacity, we combine dynamic and nonmonotonic (preferential) logics
in a modal logic setting. We extend a fragment of Van Benthem and De
{Rijke}'s dynamic modal logic with additional preferential operators in
the underlying static logic, which enables us to define defeasible
(pragmatic) entailments over a given piece of discourse. We will show
how this setting can be used for a dynamic logical analysis of
preferential resolutions of ambiguous pronouns in discourse. 
\end{abstract}

\section{Introduction} \label{sec:intro}
\index{discourse!preferences}\index{dynamic logic}

The goal of model-theoretic semantics is to establish an
interpretation function from the expressions of a given language to a
class of well-understood mathematical structures (models). This
enables a formal logical understanding of what an expression means and
what its consequences are.  For instance, natural language semantics
has recently developed a relatively simple {\em dynamic\/}
model-theoretic understanding of the interplay between
indefinite descriptions and anaphoric bindings.  
\index{dynamic semantics}
These dynamic
semantic theories of natural language give model-theoretic
explanations of {\em possible\/} anaphoric bindings, 
\index{anaphoric binding!possible}
assuming that
additional pragmatics will address the issues of anaphora resolution.
\index{anaphora resolution}
A correct dynamic semantic analysis predicts each of the possible
referents available in the context, just as a classical logical
analysis `lists' all possible scoping and lexical ambiguities.

Consider the following simple discourses~(\ref{jmb}) and (\ref{bmj}).
\enumsentence{\label{jmb} John met Bill at the station. \tts{He}{1} greeted
\tts{him}{1}.}  
\enumsentence{\label{bmj} Bill met John at the
station. \tts{He}{1} greeted \tts{him}{1}.}  
The two discourses are
semantically equivalent.  A precise dynamic semantic analysis would
treat \tts{he}{1} and \tts{him}{1} in both examples as variables that
range over the semantic values of {\sf John} and {\sf Bill}, with the
additional constraint that the referents of \tts{he}{1} and
\tts{him}{1} are different.  This analysis predicts two sets of
equally possible bindings. There is, however, a clear preferential
difference between the two discourses. There is a preference for the
bindings, \tts{he}{1} = {\sf John} and \tts{him}{1} = {\sf Bill},
in~(\ref{jmb}), and for the opposite bindings, \tts{he}{1} = {\sf
Bill} and \tts{him}{1} = {\sf John}, in~(\ref{bmj}).  
\index{anaphoric binding!preferred}

Preferential effects on discourse interpretations and the entire issue
of ambiguity resolution 
\index{ambiguity resolution}
have traditionally been put outside the scope
of logical semantics, into the more or less disjoint subfield of
`pragmatics.'  This academic focus sharply contrasts with the
importance placed on disambiguation and resolution issues in natural
language processing (or computational linguistics), 
\index{natural language processing}\index{computational linguistics}
where realistic
accounts of naturally occurring discourses and dialogues are
demanded from application systems. Computational accounts,
however, often fall short of logical or model-theoretic
formalizations. In artificial intelligence (AI), in contrast, logical
formalization of pragmatics, or defeasible reasoning, 
\index{defeasible reasoning}\index{nonmonotonic logic}
was brought into
the central focus of research at an early stage
\cite{McCarthy+Hayes:69}, and led to the development of nonmonotonic
logics. 
\index{artificial intelligence}\index{AI}

More recently, there are proposals to incorporate defeasible
reasoning within natural language semantics to approximate
the class of realistic conclusions of a given sentence or discourse
\cite{Veltman:dius,Lascarides+Asher:93}. 
In contrast with these specific proposals,% 
\footnote{\namecite{Veltman:dius} defines default reasoning
in terms of his update semantics. 
\namecite{Lascarides+Asher:93} extend Discourse Representation Theory (DRT)
with the definition of commonsense entailment given by
\namecite{AshMor:ce}.}  
we will propose a {\em
general\/} framework for
preferential dynamic semantics, 
\index{dynamic semantics!preferential}
and illustrate how the basic properties of
discourse pragmatics
\index{discourse pragmatics} 
exhibited by ambiguous pronouns can be encoded
within the framework.

The present framework combines a general model of nonmonotonic logic
\cite{Shoham:ractacftsoai} and a general model of dynamic logic
\cite{Benthem:lia,Rijke:asodml}.  
In this logical setup, we
specify defeasible information and associated entailment relations
over a given discourse, and classify the relative stability of
conclusions made on the basis of this additional defeasible
information. 
\index{defeasible information}
Our paper is about a general framework of preferential
dynamic semantics that abstracts away from numerous specific
possibilities for how to represent utterance logical forms and
discourse contexts, and how to actually compute preferences.  Since
logical formalization of discourse pragmatics is in an early stage of
development, we believe that it benefits immensely from an attempt
such as here to sort out general meta-theoretical issues from
specific accounts.

The paper is organized as follows. Section~\ref{pron} summarizes the
preferential effects on ambiguous discourse anaphoric pronouns.
\index{pronouns!discourse anaphoric}
Section~\ref{pdml} presents our basic logical framework.
Section~\ref{tapdl} illustrates formalisms at work in pronoun
interpretation in a first-order discourse logic.

\section{Preferences in Ambiguous Pronouns}\label{pron}\index{pronouns!ambiguous}

We summarize, here, the basic properties of preferential
effects on discourse semantics. 
\index{discourse semantics!preferential effects on}
We focus on ambiguous pronouns in
simple discourses, 
\index{discourse}
and illustrate the properties of dynamicity,
indeterminacy, defeasibility, and preference class interactions.

\subsection{Discourse Pragmatics as Preferential Reasoning}

Most present-day linguistic theorists assume the trichotomy of syntax,
semantics, and prag\-matics, but there is no single agreed-upon
definition of exactly what {\em linguistic pragmatics\/} is. 
\index{linguistic pragmatics}\index{pragmatics}
Some equate
it with `indexicality', some with `context dependence', and others
with `language use' \cite{Levinson:83}. There is also a common
pipeline view of the trichotomy, in that pragmatics adds
interpretations to the output of semantics that interprets the output
of syntax. In this pipeline view, the direct link between syntax and
pragmatics is lost.

We take a logic-inspired definition of pragmatics as the {\em
nonmonotonic\/} subsystem characterized by {\em defeasible\/} rules. 
\index{defeasible rules|also{preferences}}
We
also view all defeasible rules to be {\em preferences\/}, 
so the
pragmatics subsystem corresponds to a subspace of preferential
reasoning, which {\em controls\/} the subspace of {\it possible\/}
interpretations carved out by the indefeasible linguistic rules in the
`grammar' subsystem.%
\footnote{We assume, following the theoretical linguistic
tradition, that there is a linguistic rule system consisting of
indefeasible rules of morphosyntax and semantics, and call it the
`grammar subsystem'. We also assume that most commonsense rules are
defeasible, but leave the question open as to whether there are also
indefeasible commonsense rules.}  From this perspective, pragmatics is
not an underdeveloped subcomponent of semantics alone, but a system
that combines all the preferential aspects of phonology, morphology,
syntax, semantics, and epistemics. There is evidence that these
heterogeneous linguistic preferences interact with one another, and
also with nonlinguistic preferences coming from the commonsense world
knowledge. What we have then is a dichotomy of grammar and pragmatics
subsystems rather than a trichotomy. Under this view, neither
indexicality nor context dependence defines pragmatics since there are
both indefeasible and defeasible indexical and context-dependent
rules. In fact, in a {\it dynamic\/} architecture for discourse
semantics, where meaning is given to a sequence of sentences rather
than to a sentence in isolation, context dependence 
\index{context dependence}
is an inherent
architectural property supporting the anaphoricity 
\index{anaphoricity}
of
natural language expressions.

\subsection{Basic Properties of Discourse Preferences}\label{properties}

\btable
\begin{center}
\begin{tabular}{ll}\hline
\multicolumn{2}{l}{\normalsize Grammatical Effects:}\\
A. & John hit Bill. Mary told {\it him\/} to go home. \\
B. & Bill was hit by John. Mary told {\it him\/} to go
home. \\
C. & John hit Bill. Mary hit {\it him\/} too. \\
D. & John hit Bill. {\it He\/} doesn't like {\it him\/}. \\
E. & John hit Bill. {\it He\/} hit {\it him\/} back. \\
K. & Babar went to a bakery. He greeted the baker.\\
& {\it He\/} pointed at a blueberry pie. \\
L. & Babar went to a bakery. The baker greeted him.\\
& {\it He\/} pointed at a blueberry pie. \\[.1cm]
\multicolumn{2}{l}{\normalsize Commonsense Effects:}\\
F. & John hit Bill. {\it He\/} was severely injured.\\
G. & John hit Arnold Schwarzenegger. {\it He\/} was
severely injured.\\
H. & John hit the Terminator. {\it He\/} was severely
injured.\\
I. & Tommy came into the classroom. He saw Billy at the
door. \\
& He hit him on the chin. {\it He\/} was severely
injured.\\
J. & Tommy came into the classroom. He saw a group of
boys at the door.\\
& He hit one of them on the chin. {\it He\/} was severely
injured.\\\hline
\end{tabular}
\caption{Discourse Examples in the Survey}\label{deits}
\end{center}
\etable

We will now motivate four basic properties of discourse preferences
\index{preference!discourse preference}
with examples of ambiguous discourses with ambiguous
pronouns. \namecite{Kameyama:isadp} analyzed a survey result
of pronoun interpretation preferences 
\index{pronouns!interpretation preferences}
from the perspective of
interacting preference classes in a dynamic discourse processing
architecture. This analysis identified a set of basic `design
features' that characterize the preferential effects on discourse
meaning, and outlined how they combine to settle on preferred
discourse interpretations. 
\index{preferred discourse interpretations}
These basic properties can be summarized as
{\it dynamicity\/}, 
\index{dynamicity}
{\it (in)determinacy\/},
\index{(in)determinacy} 
{\it defeasibility\/},
\index{defeasibility}\index{nonmonotonicity}
and {\it preference class interactions\/}.
\index{preference class!interactions}

Table 1 shows those examples discussed by \namecite{Kameyama:isadp}. In a
survey, speakers had to pick the preferred reference of pronouns in
the last sentence of each discourse example (shown in italics).%
\footnote{The respondents were told to read the discourses with a
`neutral' intonation, for the survey was intended to investigate
only {\em unstressed\/} pronouns.\index{pronouns!unstressed}}  
Table 2 shows the survey
results.\footnote{The $\chi^2_{df=1}$ significance 
\index{$\chi^2_{df=1}$ significance}
for each example
was computed by adding an evenly divided number of the `unclear'
answers to each explicitly selected answer, reflecting the assumption
that an `unclear' answer shows a genuine ambiguity.}  These and
similar examples will be used in this paper.

\btable
\begin{center}
\begin{tabular}{llllll}
& \multicolumn{3}{l}{\normalsize Answers} & $\chi^2_{df=1}$ &
$p$\\\hline
A. & John 42 &  Bill 0 &  Unclear 5 & 37.53 & $p<.001$\\
B. & John 7 &  Bill 33 &  Unclear 7 & 14.38 & $p<.001$\\
C. & John 0 &  Bill 47 &  Unclear 0 & 47 & $p<.001$\\
D. & J. dislikes B. 42 &  B. dislikes J. 0 &  Unclear 5 &
37.53 & $p<.001$\\
E. & John hit Bill 2 &  Bill hit John 45 &  Unclear 0 &
39.34 & $p<.001$\\
K. & Babar 13 &  Baker 0 &  Unclear 0 & 13 & $p<.001$\\
L. & Babar 3 &  Baker 10 &  Unclear 0 & 3.77 &
$.05<p<.10$\\[.1cm]
F. & John 0 &  Bill 46 &  Unclear 1 & 45.02 & $p<.001$\\
G. & John 24 &  Arnold 13 &  Unclear 10 & 2.57 &
$.10<p<.20$\\
H. & John 34 &  Terminator 6 &  Unclear 7 & 16.68 &
$p<.001$\\
I. & Tommy 3 &  Billy 17 &  Unclear 1 & 9.33 &
$.001<p<.01$\\
J. & Tommy 10 &  Boy 7 &  Unclear 3 & 0.45 &
$.50<p<.70$\\\hline
\end{tabular}
\caption{Survey Results}\label{surres}
\end{center}
\etable

\subsubsection{Dynamicity}\label{ling:dyn}

We are interested in discourse pragmatics, that is, discourse semantics
enriched with preferences, so it is natural to start from where
discourse semantics leaves off, not losing what discourse semantics
has accomplished with its dynamic architecture and the view of
sentence meaning as its context change potential. We thus take {\it
dynamicity} to be a basic architectural requirement in an integrated
theory of discourse semantics and pragmatics.\footnote{There are two
levels of dynamicity that affect utterance interpretation in
discourse.  One is the utterance-by-utterance dynamicity that
affects the overall discourse meaning, and the other is the
word-by-word or constituent-by-constituent dynamicity that affects
the meaning of the utterance being interpreted.  In this paper, we
will focus on the former.}

The discourse examples (\ref{jmb}) and (\ref{bmj}), repeated here,
demonstrate the fact that the preferred interpretation of an utterance
depends on the preceding discourse context.
\enumsentence[(\ref{jmb})]{ John met Bill at the station. \tts{He}{1}
greeted \tts{him}{1}.}
\enumsentence[(\ref{bmj})]{ Bill met John at the station. \tts{He}{1}
greeted \tts{him}{1}.}
The two discourses are semantically equivalent. Two
male persons, `John' and `Bill', engage themselves
in a symmetric action of {\tt meeting}. Both individuals are available
for anaphoric reference in the next sentence, and since the two pronouns
in {\it He greeted him\/} must be disjoint in reference and each pronoun
has two possible values, dynamic semantic theories predict two equally
possible interpretations, {\tt John greeted Bill} and {\tt Bill
greeted John}. However, these discourses have different {\it
preferred values} for these pronouns. In (\ref{jmb}), due to a {\it
grammatical parallelism preference} 
\index{preference!grammatical parallelism preference}\index{parallelism}
(exhibited by discourse D in Table
1), the preferred interpretation is {\tt John greeted Bill}. In
(\ref{bmj}), the same parallelism preference leads to the reverse
interpretation of {\tt Bill greeted John}. 

Dynamic semantics has been motivated by examples such as {\it A man
walks in the park. He whistles.\/}, where an existential scope extends
beyond the syntactic sentence boundary to bind pronouns. Analogously,
preferential dynamic semantics would have to account for examples such as
\rf{jmb} and \rf{bmj}, where different syntactic configurations of the
same semantic content have different {\it extended effects\/} on the
preferred interpretation of pronouns.

\subsubsection{(In)determinacy}\label{ling:ind}

One notable feature of the survey results shown in Table 2 is that the
resulting $\chi^2_{df=1}$ significance varies widely. We consider
preference to be {\it significant\/} if $p<.05$, {\it weakly
significant\/} if $.05<p<.10$, and {\it insignificant\/} if $.10<p$ as a
straightforward application of elementary statistics.  It is
reasonable to assume that the statistical significance of a preference
corresponds to how determinate the given preference is.  Significant
preferences are thus unambiguous and determinate, and insignificant
preferences indicate ambiguities and indeterminacies.  The
preferential machinery then must allow both unambiguous and ambiguous
preferences 
\index{preference!ambiguous preference}
to be concluded, rather than always producing a single
maximally preferred conclusion.

Preferential reasoning is supposed to resolve ambiguities, however,
and unresolved preferential ambiguities make discourses incoherent. It
seems reasonable to assume a discourse pragmatic meta-principle that
says, {\it a discourse should produce a single maximally preferred
interpretation\/}. 
\index{preference!maximal preference}
Such a meta-principle is akin to Grice-style maxims
of conversation, where a preferred discourse is truthful, adequately
informative, perspicuous, relevant, and so forth \cite{Grice:75}.  It
seems that this kind of a meta-principle is needed to assure that
speakers try to avoid indeterminate preferences precisely because the
underlying preferential logical machinery does not guarantee
determinacy.

We thus identify a basic property of preferential reasoning ---
preferential conclusions are sometimes {\it determinate\/} with a single
maximally preferred interpretation, and other times {\it
indeterminate} with multiple maximally preferred interpretations. The
latter results in a genuine ambiguity, or incoherence, 
\index{incoherence}\index{ambiguity}
violating the
basic pragmatic felicity condition.
\index{pragmatic felicity condition}

Let us turn to concrete examples.  Both discourses (\ref{jmb}) and
(\ref{bmj}) have determinate preferred
interpretations due to the grammatical parallelism
preference. In contrast, discourse (\ref{jbm}) leads to no clear
preference because no relevant preferences
converge on a single determinate choice. Discourse (\ref{jbm}) is
thus infelicitous.
\enumsentence{\label{jbm} John and Bill met at the station. He greeted him.}

\subsubsection{Defeasibility}\label{ling:def}

A conclusion is {\it defeasible\/} if it may have to be retracted when
some additional facts are introduced.  This property is also called
{\it nonmonotonicity\/}, and is the defining property of {\it
preferences}. This property also defines {\it pragmatic\/}, as opposed
to grammatical, conclusions under the present assumption that
grammatical conclusions are indefeasible.

The following continuation of (\ref{jmb}) illustrates defeasibility.
\enumsentence{\label{jmbgb} John met Bill at the station. He greeted
him. John greeted him back.} 
In (\ref{jmbgb}), the third
sentence, with its indefeasible semantics associated with the adverb
{\it back\/} (as in discourse E in Table 1), forces a reversal of the
preferred interpretation concluded after the second sentence. This
on-line reversal produces a discourse-level {\it garden path\/}
effect, 
\index{garden path effect}\index{discourse!garden path effect}
analogous to the sentence-level phenomena such as in {\it The horse
passed the barn fell\/} or {\it The astronomer married a star.\/}

Garden path effects are cases of {\it preference reversal\/}, 
\index{preference!preference reversal}
which
should not be confused with explicit retractions or negations of
indefeasible conclusions. The former can be triggered implicitly,
whereas the latter must be explicitly asserted. The latter is
illustrated by the following discourse-level {\it repair\/} example,
\index{discourse!repair}
where the explicit retraction signal {\it No\/} negates the immediately
preceding assertion, and opens a way for a different fact to be
asserted in the next sentence.  
\enumsentence{\label{jmbmp} John met Bill at the
station. No. He met Paul there.}

\subsubsection{Preference Classes}\label{ling:pre}

When multiple preferences simultaneously succeed, the combined effects
are quite unlike the familiar patterns of grammatical rule
interactions. When mutually contradictory indefeasible rules both
succeed, the whole interpretation is supposed to fail. For instance,
{\it John met Mary at the station. He knows that she loves himself.\/}
leads to no indefeasible interpretation. In contrast, preferences may {\it
override} other preferences that contradict them. Ambiguities persist
only when mutually contradictory preferences are equally strong.  A
logical model of preferential reasoning, therefore, must predict
ambiguity resolutions due to overrides.

One type of override is predicted by the so-called Penguin Principle,
\index{Penguin Principle|see{defeasible inference patterns}}
where the conclusion based on a more specific premise wins (see
\namecite{Lascarides+Asher:93} for a linguistic application).  This
principle does not explain all the override phenomena in pragmatic
reasoning, however. We must posit the existence of {\it preference
classes\/} to predict overrides among groups of preferences
\cite{Kameyama:isadp}. We thus distinguish between two kinds of
conflict resolutions 
\index{conflict resolution}
in pragmatics, one due to the Penguin Principle
and the other due to preference class overrides.%
\footnote{\namecite{Asher+Lascarides:95} implement 
a class-level override in terms of a `meta-penguin principle' forced
on rule classes.  Their law of `Lexical Impotence' (p.~96) predicts
that discourse inferences generally override default lexical
inferences.} \index{preference class!overrides}
In this paper, we focus on the interaction between two
major preference classes --- the {\em syntactic preferences\/} 
\index{preference!syntactic preference}
based on
the {\it surface structure\/} of utterances%
\footnote{This includes both the
parallelism and attentional preferences discussed by
\namecite{Kameyama:isadp}.  It was conjectured there that these preference
classes may be independent subclasses of a larger `entity-level'
preference class, which is 
qualitatively different from the `propositional-level' commonsense
preference class.}  and the {\em
commonsense preferences\/} 
\index{preference!commonsense preference}
based on the {\it commonsense world knowledge\/}.
\index{commonsense world knowledge}

First consider two examples (A and B) in Table 1 repeated here.
\enumsentence{\label{jbmgh} John hit Bill. Mary told him to go home.}
\enumsentence{\label{bjmgh} Bill was hit by John. Mary told him to go
home.}
Discourses
(\ref{jbmgh}) and (\ref{bjmgh}) illustrate a syntactic preference ---
the preference for the main grammatical subject to be the antecedent
for a pronoun in the next utterance. Henceforth, this syntactic
preference is called the {\it subject antecedent preference\/}.  
\index{preference!subject antecedent preference}
In
(\ref{jbmgh}), the preferred value of the pronoun {\it him\/} is
John. In (\ref{bjmgh}), with passivization, the preferred value shifts
to Bill. Since passivization does not affect the thematic roles (such
as Agent or Theme) of these referents, we conclude that this
preference shift is directly caused by the shift in grammatical
functions.

Next, consider the following.%
\footnote{(\ref{jhb}) is a slight variation of F in Table
1. (\ref{whc}) is a variant of Len Schubert's (personal communication)
example.}
\enumsentence{\label{jhb} John hit Bill. He got injured.}
\enumsentence{\label{whc} The wall was hit by a champagne glass. It
broke into pieces.}
%also Jerry's safe combination examples?
Discourses (\ref{jhb}) and (\ref{whc}) illustrate that the above subject
antecedent preference is overridden by a stronger class of preferences
having to do with commonsense causal knowledge --- in these cases,
about hitting causing injuring or breaking.

\btable
\begin{center}
\begin{tabular}{llll|l}
   & {\normalsize Syntactic Pref.}  & {\normalsize Commonsense Pref.} & {\normalsize Semantics} &
{\normalsize Winner}\\\hline
A. & John & unclear & --- & Syntactic Pref.\\
B. & Bill & unclear & --- & Syntactic Pref.\\
C. & John & unclear & Bill & Semantics\\
D. & John--Bill & unclear & --- & Syntactic Pref.\\
E. & John--Bill & unclear & Bill--John & Semantics\\
K. & Babar & unclear & --- & Syntactic Pref.\\
L. & Baker & unclear & --- & Syntactic Pref.\\[.1cm]
F. & John & Bill & --- & Commonsense Pref.\\
G. & John & John/Arnold & --- & Commonsense Pref.\\
H. & John & John & --- & Commonsense Pref.\\
I. & Tommy & Billy & --- & Commonsense ({\tiny but difficult})\\
J. & Tommy & Boy(/Tommy) & --- & $??$\\\hline
\end{tabular}
\caption{Preference Interactions}
\end{center}
\etable

We thus assume that there are preference classes, or modules, that
independently conclude the preferred interpretation of an utterance,
and that these class-internal conclusions interact in a certain
general overriding pattern to produce the final preference.  Table 3
shows the survey result analyzed from this perspective of preference class
interactions.  Based on this analysis, we will model the
following general patterns of preference interactions:
\index{preference!preference interactions} 
\bit
\item Indefeasible syntax and semantics override all preferences.
\index{indefeasible syntax and semantics}
\item Commonsense preferences override syntactic
preferences.\footnote{This overriding can be difficult when the
syntactic preference is extremely strong. For instance, example I in
Table 1 creates an utterance-internal garden-path effect where the
first syntactically preferred choice for Tommy is retracted in favor
of a more plausible interpretation supported by commonsense
preferences.}
\item Syntactic preferences dominate the final
interpretation only if there are no relevant commonsense preferences.
\eit
The general overriding pattern we identify here is schematically shown
as follows, where $\geq$ represents a `can override' relation:
\vspace*{2mm}
\begin{center}
\begin{tabular}{|ccccc|}\hline
\begin{tabular}{c}
Indefeasible \\ Syntax and Semantics
\end{tabular}
& $\geq$ & 
\begin{tabular}{c}
Commonsense \\ Preferences 
\end{tabular}
& $\geq$ & 
\begin{tabular}{c}
`Syntactic'\\
Preferences
\end{tabular}
\\\hline
\end{tabular}
\end{center}\ \\
There are a number of questions about these preference classes. For
instance, how do they arise, how many classes are there, and why can
some classes override others?\footnote{\namecite{Kameyama:isadp} proposed
that there are three preference classes that respectively concern
preferred updates of three data structure components of the dynamic
context.  These three preference classes also seem to correspond with the three
classes of {\it discourse coherence relations\/} independently proposed
by \namecite{Kehler:95} to account for the constraints on ellipsis and
other cohesive forms. This indicates a potential integration of two
apparently unrelated notions --- dynamic context data structure components and
coherence relations.}  In this paper, we simply assume the existence
of multiple preference classes with predetermined override
relationships, and propose a logical machinery that implements their
interactions.

We will now turn to the logical machinery that will be used to model
pragmatic reasoning with the requisite properties of dynamicity,
indeterminacy, defeasibility, and preference class interactions.

\setcounter{equation}{9}
\section{Dynamic Preferential Reasoning}
\label{pdml}

We have chosen to combine dynamics and preferences in a most general
logical setting in order to achieve logical transparency and
theoretical independence in the following sense. We hope that the
logical simplicity facilitates future meta-logical investigations on
the interaction of dynamics and preferential reasoning, and 
enables applications to a wider variety of preferential (defeasible)
phenomena. We will thus combine the most general dynamic
logical approach and the most general logical approach to defeasible
reasoning we know. The dynamic (relational) setting consists of the
core of the so-called dynamic modal logic of 
Van Benthem \shortcite{Benthem:lia} and \namecite{Rijke:asodml}. Our
encoding of defeasibility follows Shoham's
\shortcite{Shoham:ractacftsoai} preferential modeling of nonmonotonic
logics.

Subsection~\ref{bdml} will outline dynamic modal logic, following Jaspars
and Krahmer's \shortcite{JasKra:ud} fragment of the original
logic.%
\footnote{To be precise, the relational part of this setting is a fragment of the relational expressivity of original dynamic logic.} 
This part encodes the dynamicity property.  Subsections \ref{pidml}
and
\ref{mp} will show how preferential reasoning can be accommodated
within this fragment of dynamic modal logic. This addition encodes
defeasibility, indeterminacy, and differentiation of preference
classes.  Finally, Subsection~\ref{rpi} discusses possible pragmatic
meta-constraints on preferential interpretation definable in this logical
setting.

\subsection{Basic Dynamic Modal Logic}\label{bdml}

\namecite{JasKra:ud} present specifications of
current dynamic semantic theories in terms of dynamic modal logic
({\sc dml}), 
\index{dynamic modal logic}\index{DML|see{dynamic modal logic}}
and show how {\sc dml} can be used as a universal setting
in which the differences and similarities among different dynamic
semantic theories can be clarified. The underlying philosophy of this
unified dynamics is that dynamic theories evolve from `dynamifying' an
ordinary logic by implementing an order of information growth over the
models of this logic. 

%\begin{figure}
%\begin{center}
%\fig{dynamification.eps}{6cm}
%\end{center}
%\caption{The `Dynamification' Process}\label{fig:dynamification}
%\end{figure}

To start with, one chooses a {\em static language\/} ${\cal L}$ to
reason about the content of {\em information states\/} $S$ by means of
an {\em interpretation function\/}: $\ldb . \rdb : {\cal L} \lora {\bf
2}^S$. 
\index{information states}
This setting most often consists of a (part of)
well-known logic interpreted over a class of well-known models.  These
models are then taken to be the units of information, that is,
information states, within the dynamic modal framework. The second
(new) step consists of a definition of an {\em order of information
growth\/}, $\sqq$, over these information states. 
\index{information growth}
We write $s \sqq t$
whenever the state $t$ contains more information than $s$ according to
this definition. The conclusive step is the choice of the dynamic
language ${\cal L}^*$, which essentially comes down to selecting
different dynamic modal operators for reasoning about the relation
$\sqq$. The triple $\langle S,\sqq,\ldb {.} \rdb \rangle$ is also
called an ${\cal L}$-{\em information model\/}.
\index{information models} 

%%% Added. This convention is needed to ensure the existence of
%%% minimal states for each nonempty set of information states (JJ).

\paragraph{Conventions.}
If $M = \langle S,\sqq,\ldb {.} \rdb \rangle$ is an ${\cal
L}$-information model, then we write $s \sqsubset t$ whenever $s \sqq
t$ and not $t \sqq s$. The state $t$ is called a {\em proper
extension\/} of $s$.  If $T \subseteq S$ then the {\em minimal states\/}
in $T$ is the set $\{t \in T \mid \forall s \in T: s \sqq t \Ra t \sqq
s\}$. 
\index{information states!minimal}
We will assume that every nonempty subset of information states
contains minimal states.  Most often, dynamic semantic theories can be
described on the basis of information models that satisfy this
constraint.

\ignore{%
\begin{our-definition} \label{def:info-model}
Let ${\cal L}$ be a language. An ${\cal L}$-{\em information model\/}
is a triple $\langle S,\sqq,\ldb .\rdb \rangle$ such that $\langle S,
\sqq \rangle$ is a preorder, and $\ldb . \rdb : {\cal L} \lora {\bf
2}^S$.
\end{our-definition}
}

\subsubsection{Static and Dynamic Meaning}\label{sadm}

On the basis of these information models, one can distinguish between
static and dynamic meanings of propositions. The {\em static
meaning\/} of a proposition $\phi \in {\cal L}$ with respect to an
${\cal L}$-information model $M = \langle S,\sqq,\ldb {.} \rdb
\rangle$, written as $\ldb \phi \rdb_M$, 
is the same as $\ldb \phi \rdb$. 
\index{static meaning|also{dynamic meaning}}
The reason is that we want to
define a dynamic modal extension ${\cal L}^*$ on top of ${\cal L}$,
which requires static interpretation as well ($\ldb . \rdb_M : {\cal
L}^* \lora {\bf 2}^S$).

Given the relational structure, that is, the preorder of information
growth $\sqq$, over the information states $S$, we are able to define
a {\em dynamic meaning\/} of a proposition. 
\index{dynamic meaning|also{static meaning}}
Roughly speaking, the
dynamic meaning of a proposition is understood as its {\em effect\/}
on a given information
state $s \in S$ .%
\footnote{Note that linguistic actions most often affect
the mental state of some chosen agents or interpreters, sharply
contrasting with physical actions that affect physical situations, as
studied in AI for analysis of so-called frame problems, e.g.,
\namecite{Shoham:ractacftsoai}.} 
In other words, we wish to define the
meaning(s) of a proposition $\phi$ {\em in the context of an information
state\/} $s \in S$: $\ldb \phi \rdb_{M,s}$.

In general, different dynamic interpretations of a proposition $\phi$
are defined according to how $\phi$ {\em operates\/} on an information
state. 
\index{information states!operations on}
For example, $\phi$ might be added to or retracted from an
information state, or, in a somewhat more complicated case, $\phi$ may
describe the content of a revision to an information state. Given such
an operation $o$, we will define the $o$-meaning of a proposition
$\phi$ with respect to an information state $s \in S$ (in $M$): $\ldb
\phi
\rdb^o_{M,s}$. The proposition $\phi$ is the {\em content\/} of an
operation and $o$ specifies the {\em type\/} of operation. In {\sc dml},
all these operations are defined in terms of the growth relation
$\sqq$.

\namecite{JasKra:ud} postulate that in most
well-known logics of mental action or change, we need only four basic
operation types: {\em extension\/} ($+$) and {\em reduction\/} ($-$),
and their minimal counterparts, {\em update\/} ($+\mu$) and {\em
downdate\/} ($-\mu$). Given an information order $\sqq$ for a given
set of information states $S$, these actions are defined as follows:
\begin{equation}
\begin{array}{rcl}
\ldb \phi \rdb_{M,s}^+&=&\{t \in S \mid s \sqq t,\ t \in \ldb \phi \rdb_M\}
\\
\ldb \phi \rdb_{M,s}^-&=&\{t \in S \mid t \sqq s,\ t \not \in \ldb \phi \rdb_M\}
\\
\ldb \phi \rdb_{M,s}^{+\mu}&=&\{t \in \ldb \phi \rdb_{M,s}^+ \mid
\A u \in S: u \in \ldb \phi \rdb_{M,s}^+ \ \& \ u \sqq t \ \Ra \ t \sqq u\} \\
\ldb \phi \rdb_{M,s}^{-\mu}&=&\{t \in \ldb \phi \rdb_{M,s}^- \mid
\A u \in S:  u \in \ldb \phi \rdb_{M,s}^- \ \& \ t \sqq u \ \Ra \ u \sqq t\}\mbox{.}
\end{array}
\end{equation}
Furthermore, for every action type $o$ we use $\ldb \phi \rdb^o_{M,T}$
as an abbreviation of the set $\bigcup_{s \in T} \ldb \phi
\rdb^o_{M,s}$ (the $o$-meaning of $\phi$ with
respect to $T$) for all $T \seq S$.  A special instance of particular importance is the
$o$-meaning with respect to the minimal states in $M$: $\min_M = \{s
\in S \mid \A t \in S: t \sqq s \Rightarrow s \sqq t\}$.  We write
$\ldb \phi \rdb^o_{M,\min}$ instead of   $\ldb \phi
\rdb^o_{M,\min_M}$,
and refer to this set as the minimal $o$-meaning of $\phi$ in $M$.
This is the meaning of a proposition with respect to an empty context.
We will also use the notation $\min_M T$ for a given subset $T \seq S$
of minimal states in $T$. 
We assumed above that $\min_S T \not = \emptyset$ whenever $T \not =
\emptyset$, and therefore, $\ldb \phi \rdb^+_{M,s} \not = \emptyset
\Rightarrow \ldb
\phi \rdb^{+\mu}_{M,s} \not = \emptyset$ (the same holds for ${-}$
with respect to ${-\mu}$).

Dynamic semantic theories most often describe relational meanings of
propositions obtained from abstractions over the context.  For
every operation $o$, we will call the relational interpretation the
$o$-{\em meaning\/} of $\phi$ (in $M$).
\begin{equation}\begin{array}{rcl}
\ldb \phi \rdb_M^o &=&
\{ \langle s,t \rangle \ | \ t \in \ldb \phi
\rdb_{M,s}^o\}\mbox{.}
\end{array}\end{equation}

Finally, a dynamic modal extension ${\cal L}^*$ of ${\cal L}$ can be
defined. It supplies unary dynamic modal operators of the form
$\left[\phi\right]^o$ and $\left\langle\phi\right\rangle^o$, whose
static interpretations are as follows:
\index{dynamic modal logic!dynamic modal operators}
\begin{equation}
\begin{array}{rcl}
\ldb \left[ \phi \right]^o \psi \rdb_M &=&
\{ s \in S \ | \
\ldb \phi \rdb_{M,s}^o \subseteq \ldb \psi \rdb_M\}
\\
\ldb \left\langle \phi \right\rangle^o \psi \rdb_M &=&
\{ s \in S \ | \
\ldb \phi \rdb_{M,s}^o \cap \ldb \psi \rdb_M \not =
\emptyset\}\mbox{.}
\end{array}
\end{equation}
For example, a proposition of the form $\upb{\phi} \psi$ means that
extending the current state with $\phi$ necessarily leads to a
$\psi$-state, while $\dodm{\phi}\psi$ means that it is possible to
retract $\phi$ from the current state in a minimal way and end up with
the information $\psi$.  In this paper, we will discuss only the
extension ($+$) and update ($+\mu$) meanings of propositions.

\paragraph{Notational conventions.}
Let $C$ be a set of connectives. Then we write ${\cal L} +
C$ for the smallest superset of ${\cal L}$ closed under the
connectives in $C$. ${\cal L} \mathrel{*} C$ denotes the smallest
superset of ${\cal L}$ closed under the connectives appearing
in ${\cal L}$ and the connectives in $C$.

\subsubsection{Static and Dynamic Entailment}\label{sade}\index{discourse!entailment}

Entailments are defined as relations between sequences of formulae and
single formulae. The former contain the {\em assumptions\/} and the
latter are the {\em conclusions\/} of the entailments.  To make
concise definitions, we also define the static and
dynamic meaning of a sequence $\phi_1,\ldots,\phi_n$, abbreviated as
$\vec{\phi}$,  in a dynamic modal language ${\cal L}^*$. 
Let $M = \langle S,\sqq,\ldb {.}
\rdb
\rangle \in {\cal M}_{\cal L}$, then
\begin{equation}
\begin{array}{rcl}
\ldb \vec{\phi} \rdb_M =  \displaystyle \bigcap_{i = 1}^n
\ldb \phi_i \rdb_M & \mbox{and} &
\ldb \vec{\phi} \rdb^o_{M} = \ldb \phi_1 \rdb^o_M \circ
\ldots \circ \ldb \phi_n \rdb^o_M \mbox{.}%
\footnotemark
\end{array}\end{equation}
\footnotetext{The operation $\circ$ stands for relational
composition. For two relations $R_1,R_2 \seq S^2$: $R_1 \circ R_2 =
\{\langle s,t \rangle \in S^2 \mid \exists u \in S: R_1(s,u) \ \& \
R_2(u,t)\}$.}%  
The former part defines the static meaning of
$\vec{\phi}$, and the latter part defines the $o$-meaning of
$\vec{\phi}$.  The $o$-meaning of $\vec{\phi}$ is
the relation of input/output pairs of consecutively $o$-executing
(expanding, updating,...)  $\phi_1$ through $\phi_n$.

We will subsequently write $\ldb \vec{\phi} \rdb_{M,s}^o$ for the set
$\{t \in S \mid \langle s,t \rangle \in \ldb
\vec{\phi} \rdb_M^o\}$ and  $\ldb \vec{\phi}
\rdb_{M,T}^o = \bigcup_{s \in T} \ldb \vec{\phi}
\rdb_{M,s}^o$ for all $s \in S$ and $T \seq S$. We will write $\ldb
\vec{\phi} \rdb_{M,\min}^o$ for  the minimal $o$-meaning 
of the sequence $\vec{\phi}$.

\begin{our-definition} \label{def:sade}
Let ${\cal M}$ be some class of ${\cal L}$-information models, and let
$\phi_1, \ldots, \phi_n, \psi$ be propositions of some dynamic modal
extension ${\cal L}^*$ of ${\cal L}$. We define the following
entailments for discourse $\phi_1,\ldots,\phi_n$ ($\vec{\phi}$):
\bit

\item $\vec{\phi}$ {\em statically entails\/} $\psi$ with respect
to ${\cal M}$  
if $\ldb \vec{\phi} \rdb_M \seq \ldb \psi \rdb_M$.
\index{static entailment}

\item $\vec{\phi}$ {\em dynamically entails\/} $\psi$
according to the operation $o$ (or $\vec{\phi}$ $o$-entails
$\psi$) with respect to ${\cal M}$ 
if $\ldb \vec{\phi} \rdb_{M,s}^o
\seq \ldb \psi \rdb_M$ for all $M \in {\cal M}$ and $s \mbox{ in }
M$.
\index{dynamic entailment}

\item $\vec{\phi}$ {\em minimally\/} $o$-{\em entails\/} $\psi$ with respect to
${\cal M}$ 
if $\ldb \vec{\phi}
\rdb_{M,\min}^o \seq \ldb \psi \rdb_M$ for all $M \in {\cal M}$. 
\eit

\noindent
We use $\vec{\phi} \models_{\cal M} \psi$, $\vec{\phi} \models^o_{\cal
M} \psi$ and $\vec{\phi} \models^{\min o}_{\cal M} \psi$ as
abbreviations for these three entailment relations, respectively.

\end{our-definition}
Note that if the modal operators $\left[\phi\right]^o$ are present
within the dynamic modal language ${\cal L}^*$, then the notion of
$o$-entailment in Definition~\ref{def:sade} boils down to the static
entailment $\models_{\cal M} \left[\phi_1\right]^o
\ldots \left[\phi_n\right]^o \psi$.

When we think of operations as updates as in the following sections,
the minimal dynamic meaning of a sequence $\phi_1,\ldots,\phi_n$ is
the same as updating the minimal states (the initial context)
consecutively with $\phi_1$ through $\phi_n$. This interpretation is
the one we will use for {\em the\/} interpretation of a discourse or
text $\vec{\phi}$. Of course, as will be the case for most pragmatic
inferences, the minimal states of an information model should not be
states of complete ignorance.  To draw the defeasible
conclusions discussed in the previous section, we need to add some
defeasible background 
information. For this purpose we need the
following notation. If $\Gamma \seq {\cal L}^*$, then we write ${\cal
M}_\Gamma$ for the subclass of models in ${\cal M}$ that supports all the
formulae in $\Gamma$: 
$\{M  = \langle S,\sqq, \ldb . \rdb \rangle \in {\cal M} \mid \ldb \gamma
\rdb_M = S \mbox{ for all } \gamma \in \Gamma\}$. The entailment
$\vec{\phi} \models_{M_\Gamma}^{\min +\mu} \psi$ covers the
interpretation of a discourse $\vec{\phi}$ in the context  or 
background knowledge of $\Gamma$.

\subsection{Simple Preferential Extensions}\label{pidml}

\namecite{Shoham:ractacftsoai} introduced preferential
reasoning into nonmonotonic logics. The central idea is to add a
preferential structure over the models of the logic chosen as the
inference mechanism. 
\index{preferential structure}
This preferential structure is most often some
partial or pre-order. A nonmonotonic inference, $\phi_1,\ldots,\phi_n
\nmm \psi$, then says that $\psi$ holds in all the maximally preferred
$\vec{\phi}$-models. 

In many nonmonotonic formalisms such as Reiter's
\shortcite{Reiter:alfdr} default logic, an additional preferential
structure of an assumption set $\vec{\phi}$ is specified by explicit {\em
default assumptions\/} $\Delta$, which are defeasible. 
\index{default logic}
The central idea
is to use `as much information from $\Delta$%
 as possible' as long as it is consistent with the strict
assumptions $\Phi$. We will also encode this maximality preference in
our definition.
\index{maximality preference}
In this paper, we use a preferential operator
$\prr$ to specify the additional defeasible information. A proposition
of the form $\prr \phi$ refers to the maximally preferred
$\phi$-states.

\subsubsection{Single Preference Classes}\label{spr}

Preferential reasoning can be accommodated within the {\sc dml}
framework by assigning an additional preferential structure to the
space of information states. There are essentially two ways to do
this.   In one method, the preferential structure is added to the static structure
over information states ($\ldb . \rdb$), and in the other method, it
is added to the
dynamic structure on these states ($\sqq$). We take
the first, simpler, option in this paper.%
\footnote{The latter, more complex, option would be a more balanced combination of dynamic
and preferential reasoning because the preferential structure is
represented at the same level of information order over which
dynamicity is defined. From this perspective, the preferential
structuring of models of a given logic that supplies a 
nonmonotonic component is analogous to dynamifying a logic
by informational structuring as described by 
\namecite{JasKra:ud}. Such investigations are left for a future
study.}

As explained in Subsection~\ref{ling:pre}, the preferential reasoning for
an\-a\-phor\-ic resolution needs to take different {\em
preference classes} into account. In Subsection~\ref{mp}, we will give {\sc
dml}-style definitions for such structures, which will be a
straightforward generalization of the following definition of a single
preference class.

\begin{our-definition} \label{def:pm}
Extension with a single preference class:
\bit
\item A {\em single preferential extension\/} $\Lp$ of the static language
${\cal L}$ is the smallest superset of ${\cal L}$ such that $\prr \phi
\in \Lp$ for all $\phi \in {\cal L}$.

\item A {\em preferential\/} ${\cal L}$-model is an information $\Lp$-model
$M = \langle S,\sqq,\ldb .\rdb \rangle$, with $\ldb .
\rdb$ representing a pair of interpretation functions
$\left\langle \ld{0} . \rdb, \ld{1} . \rdb \right\rangle$ such that 
$M_0 = \langle S,\sqq,\ld{0}.\rdb \rangle$ and $M_1 =
\langle
S,\sqq,\ld{1}.\rdb \rangle$ are ${\cal L}$-information
models, and
$\ldb \phi \rdb = \ld{0} \phi \rdb
\mbox{ and }
\ldb \prr \phi \rdb = \ld{1} \phi \rdb
$ for all $\phi \in {\cal L}$. 

\item If ${\cal M}$ is a class of ${\cal
L}$-information models, then the class of all preferential ${\cal L}$
models whose nonpreferential part ($0$) is a member of ${\cal
M}$ is called the single-preferential enrichment of ${\cal M}$.

\item If ${\cal L}^* = {\cal  L}
\mathrel{+ (\mathrel{*})} C$, then $\Lp^*$ refers to the language $\Lp
\mathrel{+ (\mathrel{*})} C$. 
\eit
\end{our-definition}

The interpretation function $\ldb . \rdb$ consists of an {\em
indefeasible part} $\ld{0} .  \rdb$ and a {\em defeasible part\/}
$\ld{1} . \rdb$.  Both parts are interpretation functions of the
static language: $^{\mbox{\scriptsize \rm 0,1}}\ldb .  \rdb: {\cal L}
\lora {\bf 2}^S$. The indefeasible part replaces the ordinary
interpretation function, while the additional defeasible part is the
`pragmatic' strengthening of this standard reading. Note that a
preferential extension gives us a set of preferred states, allowing both
determinate and indeterminate interpretations.
\index{information states!preferred}

\subsubsection{Dynamic Preferential Meaning and
Preferential Entailment}\label{dpm}

Definition (\ref{psdm}) illustrates the static and dynamic {\it preferential\/}
meaning of a sentence $\phi$ analogous to the nonpreferential definitions
presented in Subsection~\ref{sadm}. The {\em static preferential
meaning\/} of a sentence $\phi$ (in a model $M$) is written as $\lll \phi
\rrr_M$, and the `dynamic' {\em preferential meaning of\/} $\phi$ {\em with respect to a
given information state\/} ({\em context\/}) $s$ in a model $M$ is written as $\lll
\phi \rrr_{M,s}$. 
\begin{equation}\label{psdm}
\begin{array}{rcl}
\lll \phi \rrr_M & = & \ldb \prr \phi \rdb ( = {^{1}}\ldb \phi \rdb ) \\
\lll \phi \rrr_{M,s}^o & = &
\left\{ 
\begin{array}{ll}
\ldb \prr \phi \rdb^o_{M,s} & \mbox{if } \ldb \prr \phi \rdb^o_{M,s} 
\not = \emptyset \\
\ldb \phi \rdb^o_{M,s} & \mbox{otherwise.}
\end{array}
\right.
\end{array}
\end{equation}
In line with the definitions of Subsection~\ref{sadm}, we write $\lll \phi
\rrr^o_M$ for the relational abstraction of $\lll \phi \rrr^o_{M,s}$. 
Our definition of the preferential dynamic meaning of a
discourse $\phi_1,\ldots,\phi_n = \vec{\phi}$ is written as $\lll \vec{\phi}
\rrr^o_{M,s}$, and its definition  deviates from the way $\ldb \vec{\phi}
\rdb^o_{M,s}$  has been defined above
because a simple relational composition of the preferential dynamic
readings of single sentences does not give us a satisfactory
definition. The failure of normal composition in this respect can be
illustrated by the following simple abstract example.
Suppose $\vec{\phi} = \phi_1,\phi_2$ is a two-sentence discourse with
\begin{equation}\label{eq:compaspref}
\ldb \prr \phi_1 \rdb^{+\mu}_{M,a} = \{b,c\} 
\mbox{, }
\ldb \phi_2 \rdb^{+\mu}_{M,1} = \{d\}
\mbox{, }\\
\ldb \prr \phi_2 \rdb^{+\mu}_{M,1} = \emptyset 
\mbox{ and } 
\ldb \prr \phi_2 \rdb^{+\mu}_{M,2} = \{e\}
\mbox{.}
\end{equation} 
We obtain both $\langle a,d \rangle,
\langle a,e \rangle \in \lll \phi_1 \rrr_M \circ \lll \phi_2 \rrr_M$.
The second pair ($\langle a,e \rangle$) is composed of maximally
preferred readings while the first pair ($\langle a,d \rangle$) is
not. Because these two pairs are both equal members of the
composition, such a  definition of the preferential meaning of a
discourse is not satisfactory.

The two-sentence discourse in this example has four possible
readings: (1) composing the two defeasible/preferential readings, (2)
composing the indefeasible reading of one sentence and the defeasible
reading of the other sentence in two possible orders, and (3)
composing the two indefeasible readings.  As we said earlier, it is
reasonable to use as much preference as possible, which means that (1)
should be the `best' composition, the two possibilities in (2)
should be the next best, and (3) should be the `worst'. We will
encode this preferential ordering based on the amount of preferences
into the entailment definition. What about then the two possible ways
of mixing indefeasible and defeasible readings of the two sentences in
the case of (2)? A purely amount-based comparison would not
differentiate them. Are they equally preferred?

In addition to the sensitivity to the amount of overall preferences,
we hypothesize that the discourse's linear progression factor 
\index{discourse!linear progression}
also gives
rise to a preferential ordering.
\index{preferential ordering}  
We thus distinguish between the two
compositions of indefeasible and defeasible readings in (2), and assign a
higher preference to the composition in which the first sentence has
the defeasible/preferential reading rather than the indefeasible
reading. The underlying intuition is that the defeasibility of
information is inversely proportional to the flow of time. It is
harder to defeat conclusions drawn earlier in the given discourse.
This has 
to do with the fading of nonsemantic memory with
time. Earlier (semantic) conclusions tend to persist, while explicit sentence
forms fade away as discourse continues. It seems easier to distinguish
(defeasible) conclusions from recently given information than from
information given earlier.

We thus take the preferential context-sensitive reading of a discourse
$\vec{\phi} = \phi_1,\ldots,\phi_n$ to be the interpretation that
results from applying preferential rules as {\em much\/} as possible and
as {\em early\/} as possible.  This type of interpretation can be
defined on the basis of an induction on the length of discourses:
\begin{equation}\label{priority}
\begin{array}{rcll}
\ld{2k} \vec{\phi} \rdb_{M,s}^o & = & \ldb \phi_n \rdb_{M,T} & 
\mbox{and}\\
\ld{2k+1} \vec{\phi} \rdb_{M,s}^o & = & \ldb \prr \phi_n \rdb_{M,T} &
\mbox{with } T = {}^k \ldb \phi_1,\ldots,\phi_{n-1} \rdb^o_{M,s}%%\footnotemark
\mbox{.} 
\end{array}
\end{equation}
Note that $k < 2^{n-1}$ in this inductive definition.  $\ld{0} \phi_1
\rdb_{M,s}$ and $\ld{1} \phi_1 \rdb_{M,s}$ are given by the ${\cal
L}$-information model $M$.  The set of states $\ld{k} \vec{\phi}
\rdb_{M,s}^o$ is called the $o$-meaning of $\vec{\phi}$ of {\em
priority} $k$ with respect to $s$ in $M$.  In this way, we obtain
$2^n$ readings of a given discourse. The {\em preferential
$o$-meaning\/} of a discourse $\vec{\phi}$ (w.r.t. $s$ in $M$) is then
the same as the nonempty interpretation of the highest priority larger
than $0$, and if all these readings are empty, then the preferential
$o$-meaning coincides with the completely indefeasible reading of
priority $0$.
\begin{equation}\label{def:pom}
\begin{array}{rcl}
\lll \vec{\phi} \rrr^o_{M,s} & = &
\ld{k} \vec{\phi} \rdb^o_{M,s}\\
\mbox{with } k & = & \max  \left(\{i \mid \ld{i} \vec{\phi} \rdb^o_{M,s} \not =
\emptyset, 0 < i < 2^n\} \cup \{0\}\right) \mbox{.}
\end{array}\end{equation}
Note that application of this definition to example~\rf{eq:compaspref}
yields $\lll \phi_1,\phi_2 \rrr_{M,0}^{+\mu} = \{e\}$.  Definition
\rf{def:pom} leads to the following succinct definition of {\em
preferential dynamic entailment}:
\index{preferential dynamic entailment}
\begin{equation}
\begin{array}{rcl}
\phi_1,\ldots,\phi_n \nmm^o_{{\cal M}} \psi & \Lra &
\lll \phi_1,\ldots,\phi_n \rrr^o_{M,s} \seq \ldb \psi \rdb_M \\
&&
\mbox{for all } s \mbox{ in } M \mbox{, for all } M \in {\cal M} \mbox{.} 
\end{array}
\end{equation}
This definition says that for every input state of a discourse
$\vec{\phi}$, the maximally preferred readings of the discourse always
lead to $\psi$-states. We write $\vec{\phi} \nmm^{\min o}_{\cal M} \psi$
whenever $\lll \vec{\phi} \rrr^{o}_{M \min} \seq \ldb \psi \rdb_M$ for
all $M \in {\cal M}$ (minimal preferential dynamic entailment).

\ignore{%
The dynamic preferential meaning of a proposition $\phi$, written as
$\lll \phi \rrr$, also depends on the kind of operation ($o$)
performed. The underlying idea is that this action is performed with
the additional defeasible information as long as this action {\em can\/}
be performed. If this last condition is not fulfilled, then the
informative action only contains the indefeasible part.  This settles
the following definition of dynamic preferential meaning (or
preferential $o$-meaning):%
\footnote{Our definition is inspired by Gabbay's 
\shortcite{Gabbay:ibfnl} formalization
of nonmonotonic logic on the basis of Kripke semantics. The
so-called possibility operator {\sf M} of McDDoy's \shortcite{McDDoy:nml1}
initial formalization of
nonmonotonic logic is adopted explicitly in the
logic in the same way as here: $s \in
\ldb \mathop{\mbox{\sf M}} \phi \rdb_M \Lolra \ldb \phi \rdb^+_{M,s}
\not = \emptyset \Lolra s \in \ldb \upd{\phi}\top \rdb_M$. 
\namecite{Jaspars:puadl} defines a
DML-style logic that comes close to Gabbay's original formulation.}
\begin{equation}
\langle\!\langle \phi \rangle\!\rangle^o_{M,s} =
\begin{array}{ll}
\ldb \prr \phi \rdb_{M,s}^o & \mbox{if } \ldb \prr \phi
\rdb_{M,s}^o \not =
\emptyset \\ [1ex]
\ldb \phi \rdb_{M,s}^o & \mbox{otherwise.}
\end{array}
\end{equation}
The dynamic preferential meaning is then defined as $\{ \langle s,t
\rangle \ | \ t \in \lll \phi \rrr^o_{M,s}\}$ and is written as $\lll
\phi \rrr^o_M$.}

\ignore{%
In general, just as in default logic, e.g., \cite{Reiter:alfdr},
preferential entailments over a discourse uses as much defeasible
information as possible --- as long as a defeasible strengthening has
a reading in the current context.}

\ignore{%
\begin{figure}
\begin{center}
\fig{pmod.eps}{7cm}
\end{center}
{\small
\[
\begin{array}{rclcrcl}
\ld{0} \phi_1 \rdb & = & \{3,4,6,8,9,10,11,12\} &&
\ld{1} \phi_1 \rdb &=& \{4,8,10,11,12\} \\ [1ex]
\ld{0}\phi_2 \rdb &=& \{2,3,5,6,7,8,9,10\} &&
\ld{1} \phi_2 \rdb &=& \{3,6,8,9\} \\ [1ex] \hline
\end{array}
\]
\begin{equation}\label{plord}
\begin{array}{ccccccc}
\{3,10\} & < & \{3\} & < & \{8,10\} & < & \{8\} \\ [1ex]
00 \ ({\bf 0}) && 01 \ ({\bf 1}) && 10 \ ({\bf 2}) && 11 \ ({\bf 3})
\end{array}
\end{equation}
}
\caption{A Preferential Model Example}\label{fig:p-entailment}
\end{figure}}

\ignore{%
Consider the preferential model of Figure~\ref{fig:p-entailment}.
Given a discourse $\phi_1,\phi_2$, there are four different readings
with different degrees of preference as follows.
\[
\begin{array}{rclcrcl}
\ldb \prr \phi_1, \prr \phi_2 \rdb^{+\mu}_{M,1}& = &
\{8\} &&
\ldb \prr \phi_1,\phi_2 \rdb^{+\mu}_{M,1}& = & \{8,10\}
\\ [1ex]
\ldb \phi_1,\prr \phi_2 \rdb^{+\mu}_{M,1}& = & \{3\} &&
\ldb \phi_1,\phi_2 \rdb^{+\mu}_{M,1}& = & \{3,10\}
\end{array}
\]
The maximally preferred $\{8\}$-reading comes from composing the two
defeasible readings $ \prr \phi_1, \prr \phi_2$. This reading
describes the combination of preferential conclusions from this
discourse. The minimally preferred interpretation of $\phi_1,\phi_2$
is the completely indefeasible reading, $\{3,10\}$. The two other
(partially) defeasible readings appear mutually unordered at first,
but we hypothesize, in part due to an efficiency consideration, that
the $\{8,10\}$-reading is preferred over the $\{3\}$-reading. It
seems natural to assume that it is easier to defeat more recent
information during a discourse interpretation. This hypothesis
establishes a linear order over the four possible interpretations as
shown in Figure~\ref{fig:p-entailment} (\ref{plord}). We thus capture
the intuition that a preference once chosen becomes less and less
defeasible as the given discourse continues.}

\ignore{%
It is not hard to determine the total preferential order over the
discourse readings on the basis their individual defeasible parts. We
can represent each sentence reading by a $0$ or $1$ depending on
whether the indefeasible or the defeasible interpretation is used. In
the case of Figure~\ref{fig:p-entailment} above, this gives us four
numbers in the system based on the number $2$ as displayed
in~(\ref{plord}). The straight order of the size of these numbers
represent their preferential degree. We will use the ordinary decimal
representation, in~(\ref{plord}) written in boldface, of the numbers
corresponding to the discourse readings. For every discourse
$\phi_1,\ldots,\phi_n$ of length $n$, each number $k \in
\{0,\ldots,2^{n}-1\}$ has a corresponding reading in a preferential
model $M$ and state $s$ in $M$, and we will refer to it by ${}^k \ldb
\phi_1,\ldots,\phi_n \rdb^o_{M,s}$.%
\footnote{It follows that in a single preferential system, if $x_1
\ldots x_n$ is
the $0$--$1$ digit sequence corresponding to number $k$ in the system
based on number $2$, then ${}^k\ldb \phi_1,\ldots,\phi_n \rdb^o_{M,s}
= \ldb \xi_1 \phi_1, \ldots, \xi_n \phi_n \rdb^o_{M,s}$ such that
$\xi_i$ is the empty symbol if $x_i = 0$, and $\xi_i = \prr$ if $x_i =
1$.}  The preferential reading of a discourse then corresponds to the
largest number $k \in \{1,\ldots,2^n-1\}$ such that ${}^k \ldb
\phi_1,\ldots,\phi_n \rdb^o_{M,s}$ is not empty. If all these readings
are empty then the preferential reading is the same as the
indefeasible reading: $\ldb \phi_1,\ldots,\phi_n \rdb^o_{M,s}$.}

\ignore{%
\begin{our-definition}\label{def:p-entailment}
Let $\displaystyle m = \max_{0 \leq k < 2^n} ({}^k\ldb
\phi_1,\ldots,\phi_n
\rdb_{M,s}^o \not = \emptyset \vee k=0)$, then
\begin{equation} \label{def:pom}
\begin{array}{rcl}
\lll \phi_1,\ldots,\phi_n \rrr_{M,s}^o  &=&
{}^m\ldb \phi_1,\ldots,\phi_n \rdb_{M,s}^o \mbox{.}
\end{array}
\end{equation}
A proposition $\psi$ is preferentially $o$-entailed by a
discourse
$\phi_1,\ldots,\phi_n$ if \\
$\lll \phi_1,\ldots,\phi_n \rrr^o_{M,s} \subseteq \ldb
\psi \rdb_M$ for all $M \in {\cal M} \mbox{ and } s \mbox{ in } M$. We
also write $\phi_1,\ldots,\phi_n \nmm_{\cal M}^o \psi$.
\end{our-definition}}

\subsection{Multiple Preference Classes}\label{mp}

Now we turn to information models of multiple preference classes
needed for formalizing the preference interaction in pronoun
resolution, as motivated in Section~\ref{pron}.  If we assume a linear
priority order on these preference classes, then it is not hard to
generalize Definition~\ref{def:pm} of a single preference class given
in Subsection~\ref{spr}. We will assume such determinate overriding
relations among preference classes here.%
\footnote{\namecite{Kameyama:isadp} points out that this is not
always the case, but in most cases, strict linearity can be enforced
through `uniting' multiple preference classes of an equal strength
into a single one: $\ldb (\prr \cup \prr') \phi \rdb = \ldb \prr \phi
\rdb \cup \ldb \prr' \phi \rdb$.}

\begin{our-definition}\label{def:mpmod}
Extension with multiple preference classes:
\bit
\item A multiple ($m$) preferential extension $\Lpm$ of ${\cal L}$ is the
smallest superset of ${\cal L}$ such that $\pri \phi \in \Lpm$ for all
$\phi \in {\cal L}$.

\item A multiple ($m$) preferential ${\cal L}$-model is a $\Lpm$-information
model $\langle S,\sqq,\ldb .\rdb \rangle$ such that $\ldb . \rdb =
\left\langle {}^0\ldb . \rdb,\ldots,{}^m\ldb . \rdb \right\rangle$
with $M_i = \langle S,\sqq,{}^i\ldb .\rdb \rangle \in {\cal M}_{{\cal
L}}$ for all $i \in \{0,\ldots,m\}$, and $\ldb \phi \rdb = {}^0\ldb
\phi \rdb \mbox{ and } \ldb \pri \phi \rdb = {}^i\ldb \phi \rdb$ for
all $\phi \in {\cal L}$ and $i \in \{1,\ldots,m\}$.

\item The class of {\em $m$-preferential enrichments\/} of a class of ${\cal
L}$ information models ${\cal M}$ is the class of all preferential
${\cal L}$ models whose indefeasible part ($0$) is a member of ${\cal
M}$. 
\eit
\end{our-definition}

Intuitively, $\prr_i \phi$ says that the current state is a preferred
state according to the $i$-th preference class and the content
$\phi$. We use a simple generalization of the preferential
dynamic meaning given in the previous section for the singular
preference setting. For a given discourse $\vec{\phi} =
\phi_1,\ldots,\phi_n$, we define $(m+1)n$ readings and define their
associated priority in the same manner as in \rf{priority}. 
Let $k < (m+1)^{n-1}$ and  $T = \ld{k}
\phi_1,\ldots,\phi_{n-1} \rdb^o_{M,s}$. Then $\ld{(m+1)k} \vec{\phi} \rdb^o_{M,s} = 
\ldb \phi_n \rdb_{M,T}^o$ and $\ld{(m+1)k + i} \vec{\phi} \rdb^o_{M,s} = 
\ldb \pr{i} \phi_n \rdb_{M,T}^o$ for all $i \in \{1,\ldots,m\}$. The
preferential $o$-meaning of $\vec{\phi}$ with respect to a state $s$
in an information model $M$, $\lll \vec{\phi} \rrr_{M,s}^o$, is
defined in the same way as for the single preferential case
\rf{def:pom}: replace $2$ with $m+1$.

\ignore{%We assign preferential orders among the different readings a 
discourse on the basis of multiple preference classes, and let the
latest information supported by the class of highest preferential
degree dominate. The definition of dynamic preferential meaning and
entailment can be copied from Definition~\ref{def:p-entailment} with a
substitution of $m+1$ for $2$. If $k \in \{0,\ldots,(m+1)^n-1\}$, then
the meaning of ${}^k \ldb \phi_1,\ldots,\phi_n \rdb^o_{M,s}$ refers to
$\ldb \prr_{x_1} \phi_n,\ldots,\prr_{x_n} \phi_n \rdb^o_{M,s}$ with
$x_1 \ldots x_n$ being the digit sequence corresponding to $k$ in the
system based on the number $m+1$ and $\prr_0$ being the empty symbol.}

\ignore{%
\begin{our-example}
Add $\ld{2} \phi_1 \rdb = \{3,6,8\}$ and
$\ld{2} \phi_2 \rdb = \{2,9,10\}$ to the interpretations
of
Figure~\ref{fig:p-entailment}. This establishes the
following nine
interpretations:}
\ignore{%
{\scriptsize
\[
\begin{array}{c|c|c|c|c|c|c|c|c}
00 \ (\bfs{0}) & 01 \ (\bfs{1}) & 02 \ (\bfs{2}) & 10 \ (\bfs{3}) & 11
\ (\bfs{4}) & 12 \ (\bfs{5}) & 20 \ (\bfs{6}) & 21 \ (\bfs{7}) & 22 \
(\bfs{8}) \\ [1ex] \{3,10\} & \{3\} & \{9,10\} & \{8,10\} & \{8\} &
\{10\} & \{3,6\} & \{3,6\} & \emptyset
\end{array}
\]}\noindent%
Because the most defeasible reading ({\bf 8}) is empty,
the
preferential (${+\mu}$-) reading of the $\phi_1,\phi_2$
discourse with respect to information state $0$ is
$\{3,6\}$ ({\bf 7}):
\begin{equation}
\lll \phi_1,\phi_2 \rrr_{M,0}^{+\mu} =
{}^7\ldb \phi_1,\phi_2 \rdb_{M,0}^{+\mu} = \{3,6\}
\end{equation}
\end{our-example}}

\subsection{Pragmatic Meta-constraints}\label{rpi}\index{pragmatic meta-constraints}

For most applications, however, this definition is far too general,
and we need to regulate the interplay of indefeasible and defeasible
interpretations with additional constraints.  We discuss some
candidates here. Let $M = \left\langle S,\sqq,\left\langle \ld{0}
. \rdb,\ld{1} . \rdb \right\rangle \right\rangle$ be a preferential
${\cal L}$-model.

\begin{principle} ({\sl Realism}) \label{cstr1}
Every preferential $\phi$-state, or $\prr
\phi$-state, is a $\phi$-state itself:%
\footnote{Compare with the `realism' principle \cite{CohLev:iicwc}: 
all intended or goal worlds of a rational agent
should be epistemically possible. This constraint is often used to
distinguish between an agent's desires and intentions.}
\[
\ld{1} \phi \rdb \seq \ld{0} \phi \rdb \mbox{.}
\]
\end{principle}
This principle is perhaps too strict. In some types of defeasible
reasoning, we would like to assign preferential meanings to
meaningless or ill-formed input, which would give us the robustness
to recover from errors. Such robustness can be expressed in terms of a
restriction to nonempty indefeasible readings as follows: $\ld{0} \phi
\rdb\not = \emptyset \Rightarrow \ld{1} \phi \rdb \subseteq \ld{0}
\phi \rdb$ (Robust Realism).

\begin{principle} ({\sl Minimal Preference}) \label{minpref}
In minimal information states, if a proposition has an indefeasible
reading, it should also have a preferential reading:
\[
\ldb \phi \rdb_{M,\min}^o \not = \emptyset \Rightarrow 
\ldb \prr \phi \rdb_{M,\min}^o \not = \emptyset \mbox{.}
\]
\end{principle}

The intuition here is that in a minimal state there
should be no obstacles that prevent the interpreter from using
his preferential expectations or prejudices. In section~\ref{tapdl}, we
will discuss some variants of this principle, which are required to
account for certain anaphora resolution preferences.

\begin{principle} ({\sl Preservation of Equivalence})
\label{cstr2}
Two propositions with the same indefeasible content
should also have the same defeasible content:
\[
\ld{0} \phi \rdb = \ld{0} \psi \rdb \Ra \ld{1} \phi \rdb
= \ld{1}
\psi \rdb \mbox{.}
\]
\end{principle}
This principle is not always desirable.%
\footnote{This principle is often used in nonmonotonic logics. It
implies, for example, the dominance of the default conclusions from
more specific information 
%%JJ 13/8/96
($\ld{0} \phi \rdb \subseteq \ld{0} \psi
\rdb \Rightarrow \ld{1} \phi \wedge \psi \rdb = \ld{1} \phi \rdb$)
. If
$\penguin \wedge \bird$ is equivalent to $\penguin$, then
Principle~\ref{cstr2} makes all the preferential information based on
$\penguin$ applicable, while the preferential information based on
$\bird$ may be invalid for $\penguin \wedge \bird$.} 
For example, in
discourses~(\ref{jmb}) and (\ref{bmj}), {\it John met Bill\/} and
{\it Bill met John\/} have the same semantic/indefeasible content, but
different pragmatic/defeasible readings. However, some weaker types of
equivalence preservation need to play a role for a satisfactory
treatment of anaphoric resolution. Such weakenings will also be
discussed in section~\ref{tapdl}.

%If this were not the case,
%then the preferential readings of the complete discourses are equal
%(the second sentences of the two discourses are equal.)

\begin{principle} ({\sl Complete Determinacy})
\label{cstr3}
Every preferential $\phi$-extension of a given
information state $s$ has at most one maximal element.
\[
\# (\ld{1} \phi \rdb^{+\mu}_{M,s}) \leq 1 \mbox{ for all
} s \mbox{.}
\]
\end{principle}
This excludes indeterminacy described in Subsection~\ref{ling:ind},
prohibiting Nixon Diamonds. Intuitively, it says that pragmatics
always enforces certainty. In other words, in cases of semantic
uncertainty, pragmatics always enforces a single choice. For example,
discourse (\ref{jbm}) should always lead to a single pragmatic
solution. Therefore, as argued earlier, this constraint 
is also unrealistic.

%%%%%%%%%%%%Analysis, JJ Feb. 16%%%%%%%%%%%%%%

\section{Toward a Preferential Discourse Logic}\label{tapdl}

We will discuss, here, two different instances of
preferential extensions of the {\sc dml}-setting of the previous
section. As we have seen, such an instantiation requires a
specification of static and dynamic modal languages and a class of
information models.  In Subsection~\ref{sppdl}, we will discuss a
simple propositional logic, and explain how simple defeasible
(preferential) propositional entailments can or cannot be drawn from a
set of preferential rules.  Our examples will illustrate the
defeasible inference patterns 
\index{defeasible inference patterns}
commonly called the Penguin Principle
\index{Penguin Principle|see{defeasible inference patterns}}
and the Nixon Diamond.  
\index{Nixon Diamond|see{defeasible inference patterns}}
In Subsection~\ref{pdrt}, we will define a
much richer dynamic semantics that integrates the defeasible
propositional inferences explained in Subsection~\ref{sppdl} into anaphora
resolution preferences. Such a combination is needed to account for
the preferential effects on anaphora resolution.
In Subsection~\ref{focfpdr}, we will define first-order variants of
pragmatic meta-constraints. In Subsection~\ref{pddp}, we will illustrate
the first-order preferential discourse logic with discourse examples
with ambiguous pronouns as discussed in Section~\ref{pron}.

\subsection{A Simple Propositional Preferential Dynamic Logic}\label{sppdl} 

Table~\ref{def:spdl1} gives a {\sc dml}-specification of a
simple dynamic propositional logic. 
\index{dynamic propositional logic}
The single preferential extension
of this logic illustrates how preferential entailments are established
according to the definitions given in the previous section.
\btable
\begin{tabular}{lp{85mm}}\hline
Static Language (${\cal L}$): & A set of {\em literals\/}: $\Prop \cup
\{\neg p \mid p \in \Prop\}$\\ 
Dynamic Language (${\cal L}^*$): & ${\cal L} +
\{{\upbm{.}},{\updm{.}}\}$\\
States ($S$): & arbitrary nonempty set.\\ 
Order (${\sqq}$): & arbitrary preorder over $S$.\\ 
Interpretation ($\ldb . \rdb$): &
A function ${\cal L} \mapsto \wp (S)$ such that\\ &
\begin{tabular}{rl}
({\it i\/}) &
$\forall \phi \in {\cal L}, s,t \in S: s \sqq t, s \in \ldb \phi \rdb
\Rightarrow t \in \ldb \phi \rdb$.\\
({\it ii\/}) &
$\forall p \in \Prop: \ldb p \rdb \cap \ldb \neg p \rdb = \emptyset$.\\
({\it iii\/}) &
$\forall \phi \in {\cal L}: \ldb \phi \rdb \cap \min_M S = \emptyset$.
\end{tabular}\\
\hline
\end{tabular}
\caption{A Class of Propositional Information Models}
\label{def:spdl1}
\index{information models!propositional}
\etable
The information states of this model are partial truth value
assignments for the propositional atoms: an atom is either true, false,
or undefined. The information order is arbitrary, while the
interpretation function is ({\it i\/}) {\em monotonic\/}, that is,
expansions contain more atomic information and ({\it ii\/}) {\em
coherent\/}, that is, expansions contain no contradictory information,
and furthermore, there is a constraint that ({\it iii\/}) the minimal
states have empty atomic content.

Let ${\cal M}$ be the class of all single-preferential enrichments of
this class of information models subject to both Principle~\ref{cstr1}
(Realism) and Principle~\ref{minpref} (Minimal Preference) defined in
the previous section. Let $\{\bird,\penguin,\fly\} \seq \Prop$, and
let $\Gamma$ be the following set of $\Lp^*$-formulae:

{\small
\begin{equation}
\{\upbm{\prr \bird}\fly,\ 
\upbm{\prr \bird}\neg \penguin,\ 
\upbm{\prr \penguin}\neg\fly,\\ 
\upbm{\penguin}\bird\},\footnotemark
\end{equation}}%
\footnotetext{The set $\Gamma$ prescribes that `normal birds can
fly', `normal birds are not penguins', `normal penguins cannot
fly' and that `penguins are birds'.}% 
then:
\begin{equation}\label{ex:penguin}\begin{array}{rcl}
\bird \nmm^{\min {+\mu}}_{{\cal M}_\Gamma} \fly & \mbox{and} &
\bird, \penguin \nmm^{\min + \mu}_{{\cal M}_\Gamma} \neg \fly \mbox{.} 
\end{array}\end{equation}
This entailment is validated by the following
derivation for all models $N \in {\cal M}_\Gamma$:
\begin{eqnarray*}
\lll \bird \rrr_{N,\min}^{+\mu} & = &
\ld{1} \bird \rdb_{N,\min}^{+\mu} \mbox{ and}\\ 
\lll \bird,\penguin \rrr^{+\mu}_{N,\min} & = & \ld{1} \bird,\penguin
\rdb^{+\mu}_{N,\min} = \\
\ldb \bird, \prr \penguin \rdb^{+\mu}_{N,\min} & = & \ldb \prr \penguin
\rdb^{+\mu}_{N,\min} \seq \ldb \neg \fly \rdb_N .
\end{eqnarray*}
By definition of the
entailment ${\nmm^{\min +\mu}_{{\cal M}_\Gamma}}$, we obtain the
results of \rf{ex:penguin}.

Next, suppose that $\{\republican,\pacifist,\quaker\} \seq \Prop$, and
\begin{equation}
\Delta = \{\upbm{\prr \quaker} \pacifist,\ \upbm{\prr \republican} \neg
\pacifist\}\mbox{.}
\end{equation}
Here, the preferential readings of $\quaker$ and
$\republican$ contradict each other. One may expect that we get
$\quaker, \republican \nmm^{\min {+\mu}}_{{\cal M}_{\Delta}} 
\pacifist$, because the preferences of the last sentence are taken to
be weaker in the definition~\rf{def:pom}. This is not the case,
however, because it is possible that a $\republican$ cannot be a normal
$\quaker$ ($\upbm{\republican} \upbm{\prr \quaker} \bot$) or vice
versa ($\upbm{\quaker} \upbm{\prr
\republican} \bot$). 

If such preferential blocks are removed, we obtain order-sensitive entailments: 
\begin{equation}\label{eq:rq1}\begin{array}{ll}
\quaker, \republican \nmm^{\min {+\mu}}_{{\cal M}_{\Delta'}} \pacifist &
\mbox{and}\\
\republican, \quaker \nmm^{\min {+\mu}}_{{\cal M}_{\Delta'}} \neg
\pacifist & \mbox{,}
\end{array}
\end{equation}
with $\Delta'$ denoting the set:
\begin{equation}
\Delta \cup
\{\upbm{\quaker}\updm{\prr \republican} \top,\\[1ex]
\upbm{\republican}\updm{\prr \quaker} \top \}\mbox{.}\footnotemark
\end{equation}
\footnotetext{Take $\top = \upbm{p}p$.}
Let ${\cal N}$ be the class of double-preferential enrichments of the
model given in Table~\ref{def:spdl1} subject to the realism and
minimal preference principles on both classes. Let $\Delta''$ be the set
\begin{equation}
\begin{array}{ll}
\{\upbm{\pr{1} \quaker} \pacifist,\ \upbm{\pr{2} \quaker} \quaker,\\[1ex] 
\upbm{\pr{2} \republican} \neg
\pacifist\} & \bigcup
\\[1ex]
\{\upbm{\quaker}\updm{\pr{i} \republican} \top,\\[1ex]
\upbm{\republican}\updm{\pr{i} \quaker} \top \mid
i =1,2\} & \mbox{.}
\end{array}
\end{equation}
The second rule says that the $\pr{2}$-reading of $\quaker$ does not
entail any information in addition to its indefeasible reading. 
In this setting, the two variants in $\rf{eq:rq1}$ yield the same
conclusion dominated by the $\pr{2}$-reading of $\republican$:
\begin{equation}\label{eq:rq2}
\begin{array}{ll}
\quaker, \republican \nmm^{\min {+\mu}}_{{\cal M}_{\Delta''}} \neg 
\pacifist & \mbox{ and }\\
\republican, \quaker \nmm^{\min {+\mu}}_{{\cal M}_{\Delta''}} \neg
\pacifist  & \mbox{.}
\end{array}
\end{equation}

\subsection{A First-order Preferential Dynamic Semantics}\label{pdrt}

We will now come to an analysis of the discourses with ambiguous
pronouns discussed in Section~\ref{pron}.  Typical dynamic
semantic analyses of discourse, such as the relational semantics for
dynamic predicate logic \cite{GroSto:dpl} 
\index{dynamic predicate logic|also{DPL}}
or first-order DRT 
\index{discourse representation theory}\index{DRT|see{discourse representation theory}}
such as 
presented, for example, in \namecite{BenMusVis:d},%
\footnote{\namecite{JasKra:ud} discuss the DML-specification of this
semantics for DRT. On the basis of these
DML-specifications, one can transfer the present definitions of
preferential dynamic entailment to a range of dynamic
semantics.}  do not yield a satisfactory preferential dynamic
semantics when we integrate them with the preferential machinery of
the previous section.  In these types of semantic theories, dynamicity
is restricted to the value assignment of variables for interpretation
of possible anaphoric links, but to account for anaphora resolution 
we need a logic that supports a
preferential interplay of variable assignments, predicates, names, and
propositions.  In the
terminology of 
\namecite{JasKra:ud}, we need to `dynamify' more 
parameters of first-order logic than just the variable assignments.%
\footnote{\namecite{BenCep:tv} discuss such further
dynamification. \namecite{GroStoVel:cam} propose a semantic
theory that combines `propositional' and `variable' dynamics,
introducing a dynamic semantics over assignment-world
pairs. It may be possible to obtain a suitable preferential extension
of this type of semantics for our purposes as well.}  To
arrive at such extended dynamics over first-order models, we will
establish a combination of the `ordinary' dynamics-over-assignments
semantics with the models of information growth used in possible world
semantics 
\index{possible world semantics}
for classes of constructive logics.%
\footnote{See \namecite{TroDal:cim1} or \namecite{Fitting:ilmtaf} for
the case of intuitionistic logic.}

Let us first present the class of our information models. The basic linguistic
ingredients are the same as for first-order logic: $\Con$ a set of
constants, $\Var$ a disjoint countably infinite set of variables, and
for each natural number $n$ a set of $n$-ary predicates $\Pred^n$. The
static language is the same as for first-order logic except for
quantifiers and negation. The dynamic language supplies the formalism
with dynamic modal operators $\upbm{.}$ and $\updm{.}$:
\begin{equation}\label{foll}
\begin{array}{rcl}
\At & = & \{Pt_1 \ldots t_n \mid P \in \Pred^n, t_i \in \Con \cup
\Var\} \\ && \cup \{t_1 = t_2 \mid t_i \in \Con \cup \Var\}\\
{\cal L} & = & \At + \{{\wedge},{\vee},\bot\}\\
{\cal L}^* & = & {\cal L} \mathrel{*} \{\upbm{.},\updm{.}\}\mbox{.}
\end{array} 
\end{equation}

\btable
\begin{tabular}{lp{3in}}
\hline
States ($S$): & 
A collection of quadruples $s = \langle D^s, I^s_p, I^s_c, I^s_v
\rangle$ with $D^s$ a nonempty set of {\em individuals\/}, $I^s_p :
\Pred^n \lora \wp( (D^s)^n)$ the {\em local interpretation of
predicates\/}, $I^s_c : \Con \leadsto D$ a {\em partial local
interpretation of constants\/}, and $I^s_v : \Var \leadsto D$ a {\em
partial variable assignment\/}.\\ 
Order (${\sqq}$): & 
A preorder over $S$ such that \\ &
\begin{tabular}{rp{85mm}}
({\it i\/}) &
For all $s,t \in S$ if $s \sqq t$ then
$D^s \seq D^t$, $I^s_p(P) \seq I^t_p(P)$ for all predicates $P$,
$I^s_c(\mbox{\sf c}) = I^t_c(\mbox{\sf c})$ for all ${\mbox{\sf c}} \in
\Dom (I^s_c)$ and $I^s_v(x) = I^t_v(x)$ for all $x \in \Dom(I^s_v)$.\\
({\it ii\/}) &
For all $s,t \in S$ if $s \sqq t$, $d \in D^t$ and $x \in \Var \setminus
\Dom(I^s_v)$, then there exists $u \in S$ such that $s \sqq u$ and
$D^t = D^u$, $I^t_p = I^u_p$, $I^t_c = I^u_c$, $\Dom(I^u_v) =
\Dom(I^s_v) \cup \{x\}$ and $I^u_v(x) = d$.\\
({\it iii\/}) &
For all $s\in \min_M S$: $I^s_p(P)= \emptyset$ for all predicates $P$
and $\Dom(I^s_c) = \Dom(I^s_v) = \emptyset$.
\end{tabular}\\
Interpretation ($\ldb . \rdb$): & 
$\ldb Pt_1 \ldots t_n \rdb =  \{s \in S
\mid \langle I^s_t(t_1),\ldots,I^s_t(t_n) \rangle \in I^s_p(P)\}$,\\ &
$\ldb t_1 = t_2 \rdb = \{s \in S \mid I_t^s(t_1) = I_t^s(t_2)\}$,\\ &
$\ldb \phi \wedge \psi \rdb = \ldb \phi \rdb \cap \ldb \psi \rdb$,
$\ldb \phi \vee \psi \rdb = \ldb \phi \rdb \cup \ldb \psi \rdb$, $\ldb
\bot \rdb = \emptyset$.\\ \hline
\end{tabular}
\caption{A Class of First-order Information Models}\label{dismods}
\index{information models!first-order}
\etable

Table~\ref{dismods} presents the intended ${\cal L}$-information
models.  The growth of the information order $\sqq$ is subject to
three constraints. The first one ({\it i\/}) says that all the parameters of
first-order logic, that is, the domains, interpretation of predicates
and constants, and the variable assignments, grow with the information
order.  The other two constraints seem unorthodox.
%%Do you really mean `unorthodox' --- abnormal/nonstandard? 
%%JJ:> Yes, nonstandard I had in mind. Monotonicity is a normal constraint
%%JJ:> for intuitionistic logic. The other two are needed to make the
%%JJ:> growing assignments  (not in IL, but in DRT and FCS-type
%%JJ:> semantics) interact in a proper way with the growing
%%JJ:> interpretations and domains. That is, fresh variables should
%%JJ:> have a normal quantifying power.
The second constraint ({\it ii\/}) ensures the freedom of variables in
this setting. It tells us that in each state the range of a `fresh'
variable is unlimited, that is, it may have the value of each current
or `future' individual. 
\index{fresh variables}
This means that for every individual $d$ in an
extension $t$, every variable $x$ that does not yet have an assigned
value may be assigned the value $d$ in a state containing the
same information as $t$. This constraint differentiates the roles of
constants and variables in this setting. The last constraint ({\it
iii}) says that the minimal information states do not contain atomic
information.  It was also used for propositional information models in
Subsection~\ref{sppdl}.

The interpretation function is more or less standard.
Verification of an atomic sentence requires determination of all the
present terms, also for identities. 

Quantification can be defined by means of the dynamic modal
operators. For example, (\ref{msgi}) 
means that the ${\Meet}$-relation is symmetric and $\Greet$-relation is
irreflexive. 
\begin{equation}\label{msgi}
\upbm{\Meet xy}\Meet yx \mbox{ and }
\upbm{\Greet xy}\upbm{x = y} \bot \mbox{.}
\end{equation}
Ordinary universal quantification can be defined by using
identity and extension modality: $\forall x \phi = \upb{x=x}
\phi$.%
\footnote{Note that to get the proper universal reading here, we need
to be sure that $x$ is a fresh variable (e.g., in the minimal states).} 
Negation can also be defined by means of a dynamic modal
operator: $\neg \phi = \upb{\phi} \bot$.%
\footnote{A proper definition of
existential quantification does not seem feasible since $\updm{x =x } \phi$ is
not persistent. A better candidate is $\neg \forall x \neg \phi$, which
behaves persistently. For $\bot$ we may take $\upd{x=x} (x=x)$.}
A typical (singular) preferential sentence would be
\begin{equation}\label{pmg}
\upbm{\prr \Meet xy}\upbm{\prr \Greet uv} (u = x \wedge v = y) \mbox{,}
\end{equation}
which means that the concatenation of the preferential reading of a
$\Meet$ing and a $\Greet$ing pair makes the variables match according
to the grammatical parallelism preference.%
\footnote{A general implementation of the parallelism preference would require a
second-order scheme.}  

\subsection{First-order Constraints for Preferential Dynamic
Reasoning}
\label{focfpdr}

To model the preferential effects on ambiguous pronouns
discussed in Section~\ref{pron}, we need to postulate several
first-order variants of the pragmatic meta-constraints discussed in
Subsection~\ref{rpi}. 
\index{pragmatic meta-constraints!first-order variants}
The first-order expressivity of the languages
${\cal L}$ and ${\cal L}^*$ given in \rf{foll} and the fine structure
of the information models presented in Table~\ref{dismods} enable us
to calibrate these meta-constraints for preferential interpretation
on first-order discourse representations.

We will adopt only Principle~\ref{cstr1} (Realism) in its purely
propositional form.  Three other constraints that we will impose on
preferential interpretation regulate some `harmless' interplay of
preferences and terms. Let $M = \left\langle S,\sqq,\ldb . \rdb
\right\rangle$ be a preferential ${\cal L}$-model with $\ldb . \rdb =
\left\langle \ld{0} . \rdb, \ld{1} . \rdb \right\rangle$.

To begin with, fresh variables have no content, and therefore, we do
not allow them to block preferential interpretation.  In other words,
a proposition that contains only fresh variables as terms always has
a preferential $+\mu$-reading whenever it has an indefeasible
$+\mu$-meaning.  In fact, this is a variant of
Principle~\ref{minpref}, the principle of minimal preference.

\begin{principle} \label{minfresh}
({\sl Minimal Preference for Fresh Variables)} \ 
Let $s$ be an information state in an
information model of the type described in Table~\ref{dismods}. If
$\Dom(I^s_v)$ has an empty intersection with the variables occurring
in a given proposition $\phi$, and no constants
occur in $\phi$, then
\[
\ldb \phi \rdb_{M,s}^{+\mu} \not = \emptyset \Rightarrow
\ldb \prr \phi \rdb_{M,s}^{+\mu} \not = \emptyset \mbox{.}
\]
\end{principle}

The two other constraints for first-order discourses are obtained by
weakening Principle~\ref{cstr2} (Preservation of
Equivalence). Although this principle itself is too strong, we would
like to have some innocent logical transparency of the preferential
operator. We thus postulate Principles~\ref{transvar}
and \ref{transid}.

\begin{principle} \label{transvar}
({\sl Preservation under Renaming Fresh Variables.}) 
Preferential readings should be maintained
when fresh variables are replaced by other fresh variables:
\[
\forall x,y \in \Var \setminus \Dom(I^s_v):
s \in \ldb \prr \phi \rdb_{M} \Lra 
s \in \ldb \prr \phi[x/y] \rdb_{M} \mbox{.}
\]
\end{principle}

\begin{principle} \label{transid}
({\sl Preservation of Identities.}) 
Preferential readings should be insensitive to
substitutions of equals:
\[
\forall t_1,t_2 \in \Var \cup \Con:
s \in \ldb \prr \phi \wedge t_1 = t_2 \rdb_M \Lra 
s \in \ldb \prr \phi[t_1/t_2] \rdb_M \mbox{.}
\]
\end{principle}

\subsection{Preferential Dynamic Disambiguation of Pronouns}\label{pddp}

We will now account for the discourse examples with ambiguous pronouns
discussed in Section~\ref{pron} using the first-order preferential
discourse logic defined here. 

\subsubsection{Single-preferential Structure}

We will first examine the single-preferential structure of the
`John met Bill' sentences \rf{jmb}--\rf{jmbgb}. 
Assume the single-preferential extensions ${\cal M}$ of the models
presented in Table~\ref{dismods} subject to 
Principles~\ref{cstr1}, \ref{minfresh}, \ref{transvar}, and
\ref{transid}. This model, together with the background information $\Gamma$ containing \rf{msgi}
and \rf{pmg}, yields the intended defeasible
conclusions as follows:
\begin{equation}\label{ex:jmb}\begin{array}{rcl}
x = \john \wedge y = \bill,\Meet xy, \Greet uv 
& \nmm^{\min + \mu}_{{\cal M}_\Gamma} & (u = \john \wedge v =
\bill)%\footnotemark 
\\
x = \bill \wedge y = \john,\Meet xy, \Greet uv 
& \nmm^{\min + \mu}_{{\cal M}_\Gamma} & (u = \bill \wedge v = \john)\mbox{.}
\end{array}
\end{equation}
%\footnotetext{This is the {\sc dml}-translation of a DRS of the form
%\drs{x,y}{x = \john\\y=\bill\\ \Meet x,y \\ \Greet u,v} as given by
%\cite{JasKra:ud}.} 
This class also entails the invalidity of this kind of a
determinate resolution for the `John and Bill met'-case~\rf{jbm}:
\begin{equation}\label{ex:jbm}
\begin{array}{rcl}
x = \john \wedge y = \bill,\Meet xy \wedge \Meet yx,\\ 
\Greet uv 
& \not \nmm^{\min + \mu}_{{\cal M}_\Gamma} & (u = \john \wedge v = \bill)\mbox{.}
\end{array}
\end{equation}
The underlying reason is that the preferential meaning of $\Meet xy
\wedge \Meet yx$ may be different from that of $\Meet xy$ or
$\Meet yx$, although these three sentences all have the same
indefeasible meaning in ${\cal M}_\Gamma$.

For discourse~\rf{jmb} extended with the sentence {\it John
greeted back\/} in \rf{jmbgb}, the defeasible conclusion of the first
discourse in \rf{ex:jmb} will be invalid over ${\cal M}_{\Gamma}$:
\begin{equation}\label{ex:jmbgb}
\begin{array}{rcl}
x = \john \wedge y = \bill,\Meet xy, \Greet uv,\\ 
\Greet xu & \not
\nmm^{\min + \mu}_{{\cal M}_{\Gamma}} & (u = \john \wedge v = \bill)\mbox{.}
\end{array}
\end{equation}
The reason is that for every model $M \in {\cal M}_\Gamma$:
\begin{equation}
\forall s \in S: s \in \lll x = \john \wedge y = \bill,\Meet xy, \Greet uv
\rrr_{M,\min}^{+\mu} \Ra\\ 
\ldb \Greet xu \rdb_{M,s}^{+\mu} =
\emptyset \mbox{.}
\end{equation}

\subsubsection{Double-preferential Structure}

We will now illustrate how the overriding effects of commonsense
preferences illustrated in \rf{jhb} and \rf{whc} come about in a
double-preferential extension of the {\sc dml}-setting in
Table~\ref{dismods}.  In these cases, we hypothesized that the
commonsense preferences about hitting / injuring / breaking override the
syntactic preferences underlying the `John met Bill'
examples \rf{jmb}--\rf{jmbgb}. We postulate the following 
double-preferential background for the `hitting' scene:
\begin{equation}\label{pmh}
\begin{array}{ll}
\upbm{\pr{1} \Hit xy} \upbm{\pr{1} \Injured v} v = x & \mbox{ and }\\
\upbm{\pr{2} \Hit xy} \upbm{\Injured v} v = y   &  \mbox{.}
\end{array}
\end{equation}
The $\pr{2}$-class is associated with commonsense preferences with a
higher preferential rank, while the $\pr{1}$-class is associated with
`syntactic' preferences with a lower preferential rank. Note that we
take the commonsense impact of the word $\Hit$ so strongly that every
$\Injured v$-continuation --- not only the preferred readings of
this sentence --- leads to the defeasible conclusion that the hittee is
the one who must be injured.

The above double-preferential account also enables a formal
distinction among discourses F (same as \rf{jhb} involving Bill),
G (involving Schwarzenegger), and H (involving the Terminator) in
Table~\ref{deits}, whose differences are exhibited in the survey
results presented in Table~\ref{surres}.

Let ${\cal N}$ be the class of double-preferential enrichments of the
models of Table~\ref{dismods} satisfying the same principles as ${\cal
M}$ for both preference classes.  When $\Delta$ represents the set
containing the two preferential update rules given in \rf{pmh}, we
obtain a determinate preference for F:
\begin{equation}\label{eq:jhb}
x = \john \wedge y = \bill,
\Hit xy, \Injured v \nmm^{\min {+\mu}}_{{\cal N}_{\Delta}} v = \bill
\mbox{.}
\end{equation}
Let $\Delta'$ be the extension of $\Delta$ enriched with the following
additional commonsense rules, where $\schw$ denotes Schwarzenegger: 
\begin{equation}\label{phs}
\upbm{\pr{2} \Injured x} \upbm{x = \schw} \bot \mbox{.}
\end{equation}
This rule says that if something
is injured, then it is not expected to be
Schwarzenegger. We then obtain a case of indeterminacy for G:
%%Is this correct? Did you mean \not for both, hence indeterminacy? 
%%Yes. It is only meant to show how a preferential logical
%%differentation of those two cases can be made (G and H).
\begin{equation}\label{eq:jhs}
\begin{array}{l}
x = \john \wedge y = \schw,
\Hit xy, \Injured v \not \nmm^{\min {+\mu}}_{{\cal N}_{\Delta'}} v = \schw
\mbox{ and }\\
x = \john \wedge y = \schw,
\Hit xy, \Injured v \not \nmm^{\min {+\mu}}_{{\cal N}_{\Delta'}} v =
\john \mbox{.}
\end{array}
\end{equation}
Let $\Delta''$ be the union of $\Delta$ and the following additional
rules, where the constant $\cyb$ denotes the Terminator:
\begin{equation}\label{phc}
\upbm{\john = \cyb} \bot \mbox{ and } \upbm{\Injured
\cyb} \bot \mbox{.}
\end{equation}
The second sentence says that
the Terminator cannot be injured. This background information
establishes the preferred meaning of H:
\begin{equation}\label{eq:jhc}
x = \john \wedge y = \cyb,
\Hit xy, \Injured v  \nmm^{\min {+\mu}}_{{\cal N}_{\Delta''}} v = \john
\mbox{.}
\end{equation}
Substitution of $\Theta = \Delta' \cup \Delta''$ for $\Delta$ in \rf{eq:jhb}, for $\Delta'$ in
\rf{eq:jhs} and for $\Delta''$ in \rf{eq:jhc} yields the same conclusions as above. 
In summary, if $\Theta$ was our background knowledge, then the discourse F
predicts that Bill is injured, while G yields indeterminacy in its
preferential meaning. Discourse H preferentially entails that John is
injured.

%%%%%%%%%%%%%%%%%%%%%%%%%%%%%%%%%%%%%%%%%%%%%%%%%%%%%%%%%%%%%%%%%%%
%%%%%%%%%%%%%%       CONCLUSIONS
%%%%%%%%%%%%%%%%%%%%%%%%%%%%%%%%%%%%%%
%%%%%%%%%%%%%%%%%%%%%%%%%%%%%%%%%%%%%%%%%%%%%%%%%%%%%%%%%%%%%%%%%%%

\section{Conclusions and Future Prospects}

As a general logical basis for an integrated model of discourse
semantics and pragmatics, we have combined dynamics and preferential
reasoning in a dynamic modal logic setting.  This logical setting
encodes the basic discourse pragmatic properties of dynamicity,
indeterminacy, defeasibility, and preference class interactions
posited in an earlier linguistic analysis of the preferential effects
on ambiguous pronouns. It also provides a logical architecture in
which to implement a set of meta-constraints that regulates the
general interplay of defeasible and indefeasible static and dynamic
interpretation. We have given a number of such meta-constraint
candidates here. Further logical and empirical investigations are
needed before we can choose the exact set of constraints we need.

We demonstrated how a general model theory of dynamic logic can be
enriched with a preferential structure to result in a relatively
simple preferential model theory. We defined the preferential dynamic
entailments over given pieces of discourse, which predict that
preferential information is used as much as possible and as early as
possible to conclude discourse interpretations. That is, earlier
defeasible conclusions are harder to defeat than more recent ones. We
have also defined a logical machinery for predicting overriding
relationships among preference classes.  Overriding takes place when
later indefeasible information defeats earlier preferential
conclusions, or when a reading corresponding to a preference class of
a higher priority becomes empty and a lower preference class takes
over.  These preference class overrides give rise to conflict
resolutions that are not predictable from straightforward applications
of the Penguin Principle.

Although our focus is on pronoun resolution preferences in this paper,
we hope that our logical machinery is also adequate for characterizing
the conflict resolution patterns among various preferences and
preference classes relevant to a wider range of discourse phenomena.
The present perspective of preference interactions assumes that
preferences belong to different classes, or modules, and that there are
certain common conflict resolution patterns {\em within\/} each class
and {\em across\/} different classes.  Class-internal preference
interactions yield either determinate or indeterminate
preferences. Class-external preference interactions are dictated by
certain preexisting class-level overriding relations, according to
which the conflicts among the respective conclusions coming from each
preference class are either resolved (by class-level overrides),
ending up with the preferential conclusions of the highest preference
class (whether it is determinate or indeterminate), or unresolved,
leading to mixed-class preferential ambiguities. We would like to
investigate the applicability of this perspective to a wider range of
discourse phenomena.\footnote{It is encouraging that the recent spread
of Optimality Theory from phonology
\cite{Prince+Smolensky:93} to syntax (e.g., MIT Workshop on
Optimality in Syntax, 1995) seems to indicate the descriptive adequacy
of the fundamental preference interaction scheme, where potentially
conflicting defeasible conclusions compete for the `maximal
harmony.'} 

The present logical characterization of preferential dynamics may be
extended and/or revised in two major areas. One is the application of
actions other than updates, $+\mu$.  For example, discourse-level repairs
as in~(\ref{jmbmp}) also require reductions, $-$, and/or downdates,
$-\mu$. The other is the relational definition of preferences on the
basis of an additional structuring of the information order $\sqq$
instead of the static interpretation function $\ldb . \rdb$.  Such an
alternative definition would enable us to implement `graded'
preferences
\cite{Delgrande:aatdrbofocl}, that is, every state gets a certain
preferential status with respect to a proposition.  
\index{preference!graded}
In our paper, states were simply
declared to be preferential or nonpreferential with respect to a
proposition. Graded preferences may be required
for fine-tuning and coordinating the overall discourse pragmatics. A
question related to this topic is whether the use of graded
preferences would make the setting of multiple preference classes
superfluous.

We might also be able to extend the framework to cover on-line
sentence processing pragmatics, where the word-by-word or
constituent-by-constituent dynamicity affects the meaning of the
utterance being interpreted.  The utterance-internal garden path and
repair phenomena will then be treated analogously to the
discourse-level counterparts.
 
%%%%%%%%%%%%%%%%%%%%%%%%%%%%%%%%%%%%%%%%%%%%%%%%%%%%%%%%%%%%%%%%%%%%%%
%\bibliography{pdml}

\end{document}